\documentclass[11pt]{article}

\usepackage{acl}

\usepackage{times}
\usepackage{latexsym}
\usepackage{tabularx}
\usepackage{hyperref}
\usepackage{amsmath, amssymb}
\usepackage{amsthm}
\usepackage{stmaryrd} 
\usepackage{mathrsfs} 
\usepackage{booktabs}
\usepackage{placeins}
\usepackage{multirow}

\usepackage{xcolor}
\definecolor{upgreen}{RGB}{34,139,34}   
\definecolor{downred}{RGB}{200,30,30}   

\newcommand{\up}{\textcolor{upgreen}{\ensuremath{\uparrow}}}
\newcommand{\down}{\textcolor{downred}{\ensuremath{\downarrow}}}

\usepackage{mdframed}
\usepackage{xcolor}
\usepackage{enumitem}

\usepackage{enumitem} 
\usepackage[T1]{fontenc}

\usepackage[utf8]{inputenc}

\usepackage{microtype}

\usepackage{inconsolata}

\usepackage{graphicx}

\title{Compartmentalised Agentic Reasoning for Clinical NLI}

\author{
 \textbf{Maël Jullien\textsuperscript{1}\textsuperscript{,3}, Lei Xu\textsuperscript{3}\textsuperscript{,5}, Marco Valentino\textsuperscript{4},  Andr\'e Freitas\textsuperscript{1}\textsuperscript{,2}\textsuperscript{,3}}\\ 
 $^{1}$Department of Computer Science, University of Manchester, UK \\ 
$^{2}$ National Biomarker Centre, CRUK-MI, University of Manchester, UK\\
$^{3}$Idiap Research Institute, Switzerland \\
$^{4}$ School of Computer Science, University of Sheffield, UK\\
$^{5}$ École Polytechnique Fédérale de Lausanne (EPFL), Switzerland\\
$^{3}${\tt \{firstname.surname\}\tt@idiap.ch}
} 

\begin{document}
\maketitle

\begin{abstract}
Large language models can produce fluent judgments for clinical natural language inference, yet they frequently fail when the decision requires the correct inferential schema rather than surface matching. We introduce \textit{CARENLI}, a compartmentalised agentic framework that routes each premise–statement pair to a reasoning family and then applies a specialised solver with explicit verification and targeted refinement. We evaluate on an expanded CTNLI benchmark of 200 instances spanning four reasoning families: \textsc{Causal Attribution}, \textsc{Compositional Grounding}, \textsc{Epistemic Verification}, and \textsc{Risk State Abstraction}. Across four contemporary backbone models, \textit{CARENLI} improves mean accuracy from about 23\% with direct prompting to about 57\%, a gain of roughly 34 points, with the largest benefits on structurally demanding reasoning types. These results support compartmentalisation plus verification as a practical route to more reliable and auditable clinical inference.
\end{abstract}

\section{Introduction}
\begin{figure*}[h]
    \centering
    \includegraphics[width=\textwidth]{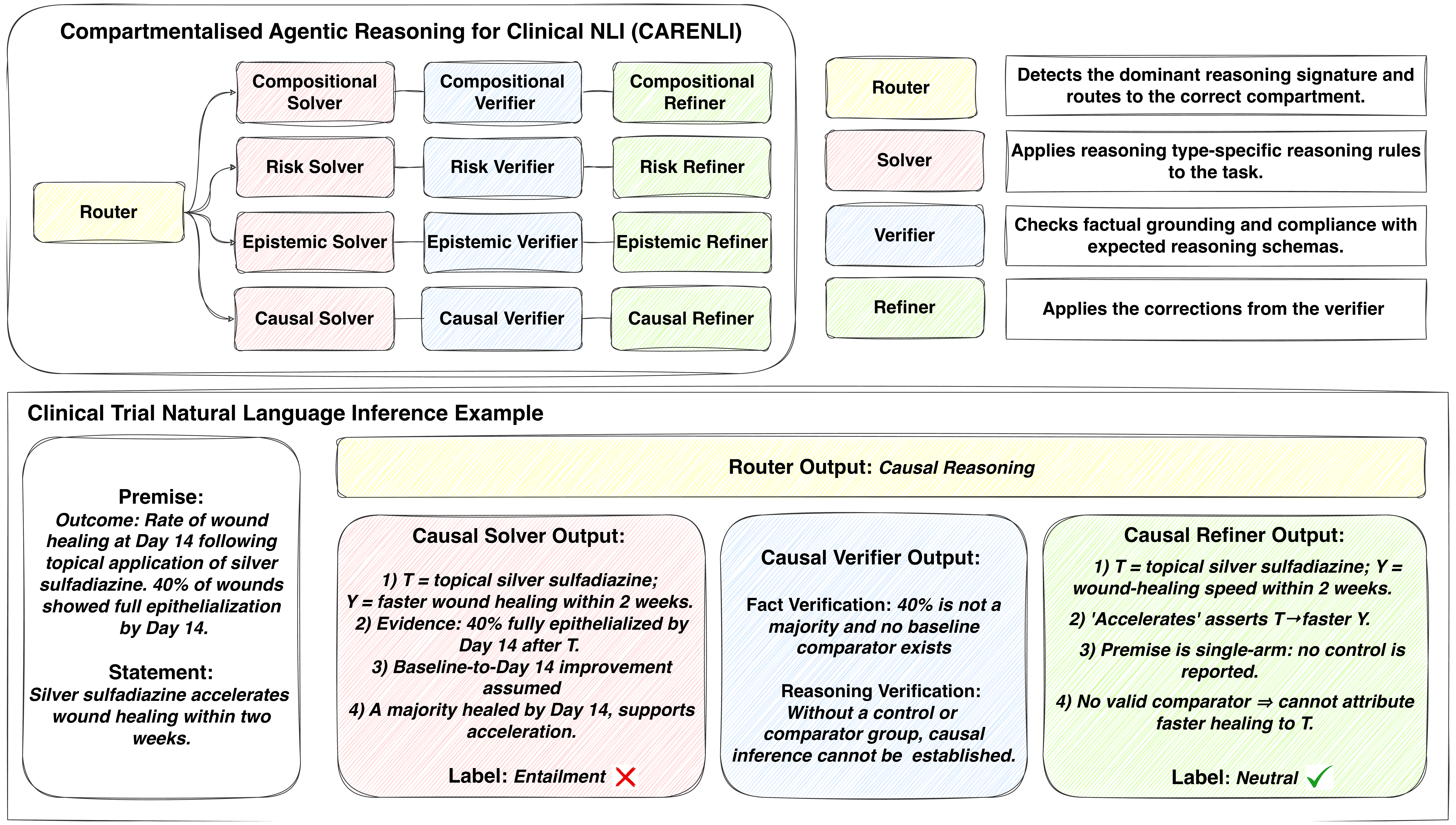}
\caption{\textit{CARENLI: Compartmentalised agentic reasoning framework for Clinical Trial NLI.} A \emph{Router} assigns each premise--statement pair to a dominant reasoning type (Causal Attribution, Compositional Grounding, Epistemic Verification, or Risk State Abstraction). A reasoning-type-specific \emph{solver} produces a provisional NLI label and explicit reasoning trace, which a \emph{verifier} audits for factual grounding and schema compliance, and a \emph{refiner} minimally corrects when needed. The figure illustrates the pipeline on a causal attribution example. Unlike generic prompting strategies \textit{CARENLI} enforces formalised, structured reasoning trajectories that are explicitly grounded in clinical trial semantics and checked for logical consistency.}
    \label{fig:main}
\end{figure*}

Despite impressive surface-level performance on natural language inference (NLI) benchmarks \cite{bowman2015snli,williams2018mnli,wang2019glue}, large language models (LLMs) exhibit systematic failures in clinical reasoning when inference requires adherence to structured, domain-specific constraints, \cite{gururangan2018annotation,mccoy2019right,ribeiro2020beyond,marcus2020rebooting,agrawal2025evaluation, jullien2023semeval, jullien2024semeval}; in settings where tolerance for systematic error is effectively zero, such failures have been associated with unsafe medical guidance and documented patient harm \cite{draelos2025unsafe,eichenberger2025bromism,reddy2025preventing, omiye2023race,goh2024diagnostic, amodei2016concrete}.

This limitation is captured by the diagnostic CTNLI benchmark of \citet{jullien2025knowledge}, which targets four core clinical reasoning capabilities: Causal Attribution, Compositional Grounding, Epistemic Verification, and Risk State Abstraction. Evaluation of six frontier large language models reveals systematic and reproducible violations of fundamental inference principles, resulting in a collapse of reasoning trajectories. Under a zero-tolerance evaluation regime, models achieve an average accuracy of only 0.25\%. Even in this constrained setting, such performance demonstrates a fundamental misalignment between current LLM reasoning mechanisms and the requirements of clinical inference, necessitating substantive corrective intervention rather than incremental optimization.

This paper tests the hypothesis that CTNLI failures are driven by schema collapse, and that enforcing schema-conditioned, compartmentalised decision procedures can mitigate this failure. We propose \emph{Compartmentalised Agentic Reasoning for Clinical NLI (CARENLI)} (Fig.~\ref{fig:main}), a direct intervention that decomposes inference into auditable roles: a \emph{Router} selects the reasoning family, a family-specific \emph{Solver} executes the corresponding procedure and emits an explicit trace, a \emph{Verifier} checks premise grounding and procedural compliance, and a \emph{Refiner} applies minimal corrections when the trace is partially correct but invalid. By explicitly binding inference to the appropriate schema, \textit{CARENLI} enforces validity criteria that reasoning-agnostic prompting leaves implicit.

Our key contributions are:
\paragraph{Phenomenon.} We characterise \emph{schema collapse} as a dominant reasoning failure mode in CTNLI, in which models reuse a generic inference pattern across heterogeneous clinical reasoning demands, relying on surface-level pattern matching and producing fluent but constraint-violating reasoning trajectories.

\paragraph{Mechanism.} We introduce \textit{CARENLI}, a compartmentalised agentic framework that performs reasoning-family routing, executes family-specific solver procedures with explicit traces, and verifies trace-level validity against the premise.

\paragraph{Validation.} We extend and expert-validate a CTNLI benchmark, and demonstrate performance gains with \textit{CARENLI} across four LLMs, with average improvements of approximately $+34$ percentage points in accuracy over generic prompting. We further analyse the contributions of \textit{CARENLI}’s components and characterise cross–reasoning-type behaviour.

\section{Methodology}
\subsection{Tasks and Dataset}
\label{sec:dataset}

This evaluation is grounded in the domain of \textit{clinical natural language inference} (CTNLI), where the objective is to determine whether a candidate \textit{statement} is entailed, contradicted, or left undetermined by a given \textit{premise}. We adopt and extend the controlled CTNLI benchmark introduced by \citet{jullien2025knowledge}, which decomposes inference into four reasoning families, each defined by explicit inferential criteria:

\begin{enumerate}[label=\textit{\roman*}.]
    \item \textsc{Causal Attribution:} distinguishes observational associations from causal claims, requiring assessment of temporality, comparators, and confounding.
    \item \textsc{Compositional Grounding:} evaluates whether clinical validity follows from the joint configuration of multiple interacting variables, such as dose, drug and schedule.
    \item \textsc{Epistemic Verification:} tests whether a claim is supported by admissible evidence rather than authority, assertion, or unsupported diagnosis.
    \item \textsc{Risk State Abstraction:} examines reasoning over latent or implicit clinical risks, particularly when severity and likelihood must be integrated.
\end{enumerate}

Each reasoning type is formally specified through typed templates that constrain the inferential operations required for a correct decision. For instance, causal attribution items require explicit evaluation of comparator conditions, while compositional grounding items require compatibility checking across multiple clinical factors. All instances are instantiated from parametric templates, ensuring that variation is restricted to clinically meaningful dimensions, such as patient characteristics, or treatment regimens, while the underlying reasoning structure remains invariant.

The benchmark has been substantially extended from its original formulation, using the same data construction and validation protocol as \citet{jullien2025knowledge}. This extended version comprises 200 instances, evenly distributed across reasoning types.

Details of the data generation pipeline, expert validation procedure, and inter-annotator agreement analysis are provided in Appendix~\ref{sec:dataset_validation}.

\subsection{Compartmentalised Reasoning}

Prior work on the CTNLI benchmark \citep{jullien2025knowledge} reveals a systematic dissociation between knowledge access and inference in large language models. Across all four reasoning families, models reliably encode the clinically relevant facts, as evidenced by near-ceiling performance on ground-knowledge probes, yet fail to apply these facts in accordance with the inferential criteria defining each task. The resulting errors are not stochastic. Instead, they are highly consistent within each reasoning type, indicating the repeated application of uniform heuristics rather than principled inference. Causal claims are reduced to temporal association, compositional validity to single-attribute matching, epistemic judgment to deference to assertion, and latent risk assessment to frequency-based reasoning. This pattern indicates a failure of inference deployment rather than knowledge acquisition.

These failures can be understood as a consequence of treating semantically heterogeneous inference problems as instances of a single undifferentiated reasoning process. The four CTNLI families correspond to distinct inferential schemas with non-interchangeable validity conditions. Causal attribution depends on counterfactual and intervention-based reasoning, compositional grounding on multi-factor compatibility constraints, epistemic verification on evidential hierarchies, and risk state abstraction on the integration of severity and likelihood. Both cognitive theories of reasoning and formal semantic analyses characterise these schemas as structurally distinct, each governed by its own representational requirements and decision rules \citep{sloman2009causal, johnson1983mental, pearl2009causality}. Collapsing these type-distinct relations into a single heuristic mode erases the constraints that make the inferences meaningful, yielding locally coherent but globally invalid judgments.

This diagnosis motivates the need for compartmentalised reasoning in clinical NLI. If errors arise because models apply a uniform inference strategy to tasks with incompatible structural demands, then improving fidelity requires enforcing explicit separation between reasoning families and constraining inference to the principles specific to each. Reasoning-agnostic prompting, including free-form Chain-of-Thought (CoT), provides no such separation and therefore permits heuristic shortcuts that systematically violate reasoning-type-specific criteria. By contrast, a compartmentalised design reinstates the boundaries between causal, compositional, epistemic, and risk-based inference, ensuring that each problem is solved using the appropriate inferential schema.

In the following section, we operationalise this principle through \emph{Compartmentalised Agentic Reasoning for Clinical NLI (CARENLI)} (Fig~\ref{fig:main}). The framework enforces compartmentalisation by decomposing inference into specialised roles, enabling auditable, schema-aligned reasoning while preserving coordination across stages.

\subsection{CARENLI}

Compartmentalised Agentic Reasoning for Clinical NLI (\textit{CARENLI}) instantiates compartmentalised reasoning through an agentic pipeline that enforces separation between problem recognition, inference execution, validation, and correction (Fig~\ref{fig:main}).

Formally, given a premise–statement pair $(p, s)$, \textit{CARENLI} computes a final entailment judgment $\Bar{y}$ via a sequence of specialised agents:
\[
(p, s)
\xrightarrow{\mathcal{R}} F
\xrightarrow{\mathcal{S}_F} (y, \tau)
\xrightarrow{\mathcal{V}_F} (v, c_f, c_p)
\xrightarrow{\mathcal{R}_F} (\Bar{y}, \Bar{\tau}).
\]
The router $\mathcal{R}$ assigns the instance to a reasoning type $F$. A type-specific solver $\mathcal{S}_F$ produces a provisional label $y$ with an explicit reasoning trace $\tau$. The verifier $\mathcal{V}_F$ audits this trace for factual grounding and schema compliance, emitting a violation signal $v$ and structured critiques $(c_f, c_p)$. When required, the refiner $\mathcal{R}_F$ applies these critiques through minimal edits to yield the final judgment $\Bar{y}$ and trace $\Bar{\tau}$ (illustrated in Fig.~\ref{fig:main})

\paragraph{Router. $(p, s)
\xrightarrow{\mathcal{R}} F$}
The Router $\mathcal{R}$ implements compartmentalisation at the level of problem recognition. It maps each $(p, s)$ to exactly one reasoning type $F \in \mathcal{F}$, where $\mathcal{F} = \{$\textsc{Causal Attribution}, \textsc{Compositional Grounding}, \textsc{Epistemic Verification}, \textsc{Risk State Abstraction}$\}$. 

\paragraph{Solver. $(p, s)
\xrightarrow{\mathcal{S}_F} (y, \tau)$}
The solver $\mathcal{S}_F$ performs reasoning-type-specific inference. Conditioned on the selected reasoning type $F$, it applies the corresponding decision procedure, and reasoning principles to produce a provisional entailment label $y$ and an explicit reasoning trace $\tau$. Each solver is constrained by the normative criteria of its reasoning type, such as comparator requirements in \textsc{Causal Attribution}, multi-factor compatibility in \textsc{Compositional Grounding}, evidential hierarchies in \textsc{Epistemic Verification}, or severity–likelihood integration in \textsc{Risk State Abstraction}. Full specifications of the reasoning-type-specific solvers are provided in Section \ref{sec:solvers}.

\paragraph{Verifier. $(y, \tau)\xrightarrow{\mathcal{V}_F}(v, c_f, c_p)$}
Verification improves LLM reliability by surfacing unsupported claims and schema violations \citep{webson2022prompt, quan2024enhancing}. Given ($y$, $\tau$), $\mathcal{V}_F$ performs fact verification, identifying non–premise-grounded or clinically inadmissible claims \citep{gravel2023learning, aljamaan2024reference}, and pattern verification, checking compliance with the reasoning-type-specific inferential schema. It outputs a violation signal $v$ and structured fact- and pattern-level critiques $(c_f, c_p)$ for downstream correction.

\paragraph{Refiner. $(y, \tau, v, c_f, c_p)\xrightarrow{\mathcal{R}_F}(\Bar{y}, \Bar{\tau})$}
Conditioned on the violation signal $v$, the refiner $\mathcal{R}_F$ applies the verifier’s critiques $(c_f, c_p)$ via minimal, schema-preserving edits to the solver’s reasoning trace $\tau$, producing a final trace $\Bar{\tau}$ and judgment $\Bar{y}$.

\section{Empirical Evaluation}

We evaluate the \emph{CARENLI} framework across four contemporary large language models: GPT-5.1 \citep{openai2025gpt5systemcard}, GPT-4.1 and GPT-4o-mini \citep{hurst2024gpt}, and DeepSeek R1 \citep{guo2025deepseek}.

Three evaluation settings are considered. First, the \emph{CARENLI} framework, in which all agents are active. Second, an \textit{Oracle \emph{CARENLI}} ablation, in which the planning stage is bypassed and the correct reasoning type is supplied directly to downstream agents and serves as a diagnostic upper bound on planning quality. This condition is not directly comparable as a deployable method, but isolates the effect of planning accuracy. Third, baseline prompting conditions, in which models are tested under CoT prompting and direct answering (Agnostic CoT, and Agnostic Direct). this setting, prompts are reasoning-type agnostic and provide no guidance toward the reasoning-type-specific inferential schemas. 

Each model–strategy pair is evaluated over the full dataset (Section~\ref{sec:dataset}) across 4 runs; the Oracle \textit{CARENLI} ablation is run once.

\section{Results}

\begin{figure*}
    \centering
    \includegraphics[width=\textwidth]{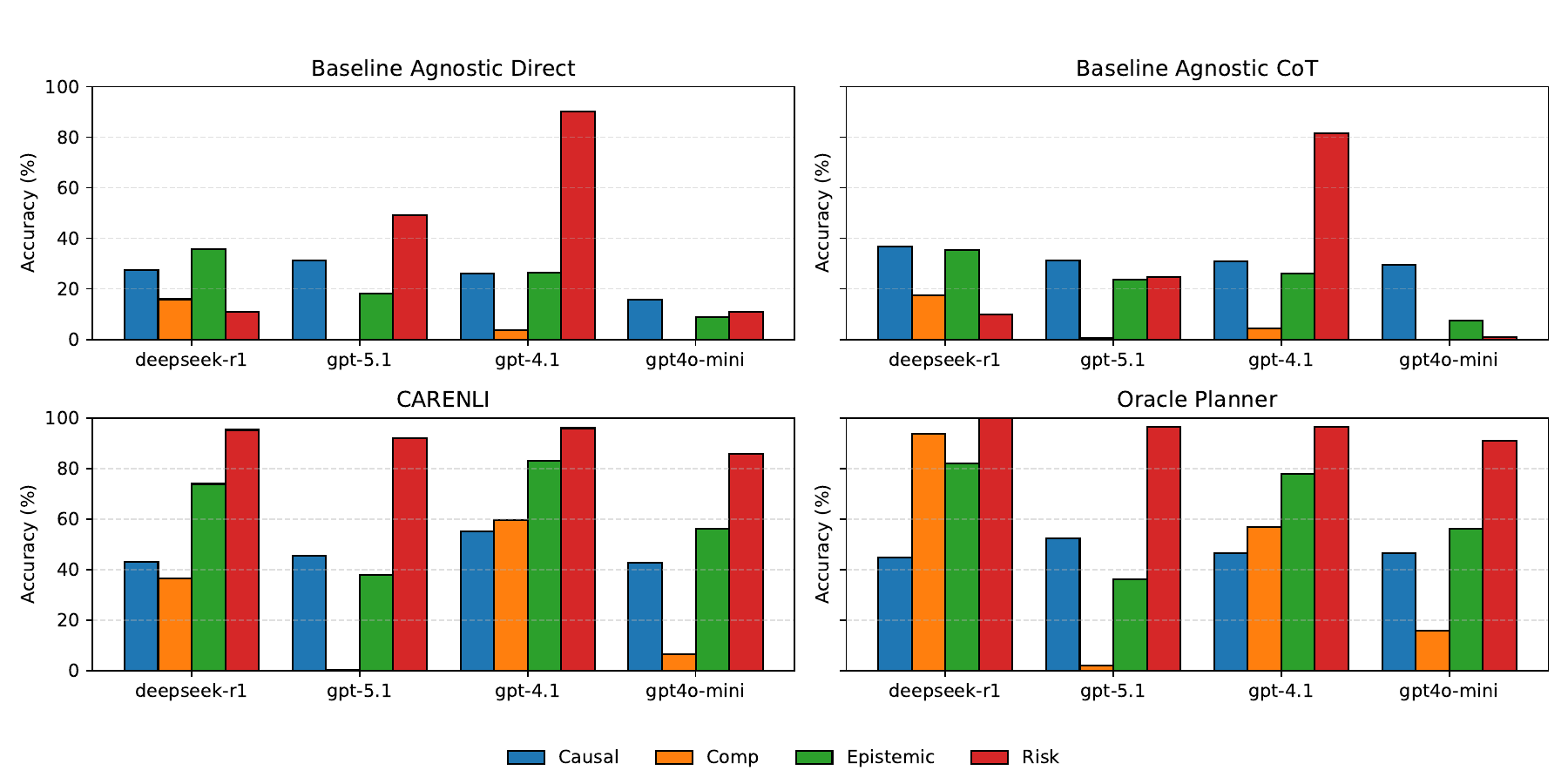}
    \caption{Overall accuracy on CTNLI tasks across all models and evaluation strategies (\emph{CARENLI}, \textit{Oracle Router}, \textit{CoT}, and \textit{Direct}). Results are averaged over four runs per configuration.}
    \label{fig:strategy_accuracy}
\end{figure*}

Figs~\ref{fig:strategy_accuracy}--\ref{fig:verifier_accuracy} and Tables~\ref{tab:routing_correct_wrong}---\ref{tab:label_reasoning} show that \textit{CARENLI} substantially improves CTNLI performance by enforcing reasoning-type-specific decision procedures rather than reasoning-agnostic prompting. Aggregated across models and task families, \textit{CARENLI} reaches 56.8\% macro accuracy, compared to 22.5\% for agnostic CoT and 23.1\% for agnostic Direct, an absolute gain of roughly +34\%. Improvements are consistent across all four backbones, though heterogeneous in magnitude (e.g., GPT-5.1 rises from 24\% to 44\%), reinforcing that scale alone does not guarantee schema-faithful clinical reasoning.

Gains are largest for \textsc{Risk State Abstraction} and \textsc{Epistemic Verification}, where \textit{CARENLI} attains 92.2\% and 62.7\% average accuracy and recovers from near-collapse baselines (Risk improves by +51.9\% relative to the strongest agnostic condition). \textsc{Causal Attribution} improves more modestly to 46.5\%, while \textsc{Compositional Grounding} remains the weakest regime despite improvement to 25.6\%, indicating persistent compositional failures in some backbones. Component analyses attribute most of the gain to the schema-constrained solver (\emph{Oracle Solver}: 60.5\%), with routing as the main bottleneck (70.8\% when correctly routed vs.\ 21.1\% when misrouted). Verifier and refiner stages contribute very marginal, task-dependent gains (overall refinement +0.7\%), primarily when solver trajectories are partially correct but procedurally incomplete.

Qualitatively, \textit{CARENLI} shifts model behavior from plausibility narratives toward schema-aligned traces that explicitly instantiate the reasoning type’s decision criteria, making typical agnostic failures traceable to specific missing checks (e.g., comparator enforcement, cross-factor constraints, or severity-weighted risk integration).

\subsection{Consistency Across Backbone Models}
\textbf{Schema-constrained prompting consistently outperforms reasoning-agnostic baselines by an average of 34\% across diverse backbones.}
As shown in Fig~\ref{fig:strategy_accuracy} and Table~\ref{tab:label_reasoning}, \textit{CARENLI} outperforms the stronger of the two agnostic prompting baselines for every evaluated model, yielding macro accuracy improvements of 37.3\% for deepseek-r1, 36.8\% for gpt-4.1, 19.4\% for gpt-5.1, and 38.2\% for gpt4o-mini. While the magnitude of improvement varies with model capacity, no backbone exhibits a regression relative to prompting baselines.

\textbf{Model scale alone is an unreliable predictor of clinical reasoning accuracy.}
The smallest improvement is observed for gpt-5.1, from 24.5\% to 43.9\%. The limited improvement observed for gpt-5.1 is consistent with prior findings that increased model scale does not reliably translate into improved clinical reasoning performance \citep{jullien2025knowledge}.

\subsection{Performance by Reasoning Type}
\textbf{\textit{CARENLI} amplifies reasoning competence where inferential structure is tractable, but leaves residual failure modes in structure-intensive regimes.} 
Fig~\ref{fig:strategy_accuracy} and Table~\ref{tab:label_reasoning} enable a reasoning-type–level analysis of \textit{CARENLI}’s effectiveness. Averaged across models, \textit{CARENLI} achieves 92.2\% accuracy on \textsc{Risk Abstraction}, 62.7\% on \textsc{Epistemic Verification}, 46.5\% on \textsc{Causal Attribution}, and 25.6\% on \textsc{Compositional Grounding}.

Relative to the strongest agnostic baseline (CoT or Direct prompting, averaged across models), these results correspond to absolute gains of +51.9\% on \textsc{Risk Abstraction}, +40\% on \textsc{Epistemic Verification}, +15\% on \textsc{Causal Attribution}, and +20\% on \textsc{Compositional Grounding}.

These gains are uneven across reasoning regimes and models. In some settings, \textit{CARENLI} recovers performance from near-baseline failure. For example, \textsc{Compositional Grounding} accuracy for GPT-4.1 increases from 3.5\% under agnostic Direct prompting to 59.5\% under \textit{CARENLI}. In contrast, for gpt-5.1, \textit{CARENLI} yields no improvement on the same reasoning type, with accuracy remaining at 0.40\%.

When reasoning families are ranked by average agnostic baseline accuracy and by average \textit{CARENLI} accuracy (Fig~\ref{fig:strategy_accuracy} and Table~\ref{tab:label_reasoning}), the ordering is identical: \textsc{Risk Abstraction}, \textsc{Epistemic Verification},  \textsc{Causal Attribution}, \textsc{Compositional Grounding}. In addition, absolute \textit{CARENLI} gains exhibit a positive correlation with baseline agnostic performance (Pearson’s $r \approx 0.54$), indicating that reasoning regimes with higher initial accuracy under agnostic prompting tend to benefit more from \textit{CARENLI}.

\textbf{\textit{CARENLI} amplifies existing latent competence rather than creating new reasoning capability.}
These results indicate that \textit{CARENLI} amplifies latent reasoning competence rather than transforming underlying reasoning capacity. Structured prompting yields substantial gains where partial inferential structure is already expressed under agnostic prompting, but does not invert task difficulty or induce robust performance in regimes characterized by near-zero baseline accuracy. \textit{CARENLI} primarily improves execution and constraint adherence, rather than introducing new compositional reasoning abilities where they are largely absent.

\subsection{Component-wise Analysis}

\subsubsection{Router}
\label{sec:planner-res}
\begin{figure}[h]
    \centering
    \includegraphics[width=\columnwidth]{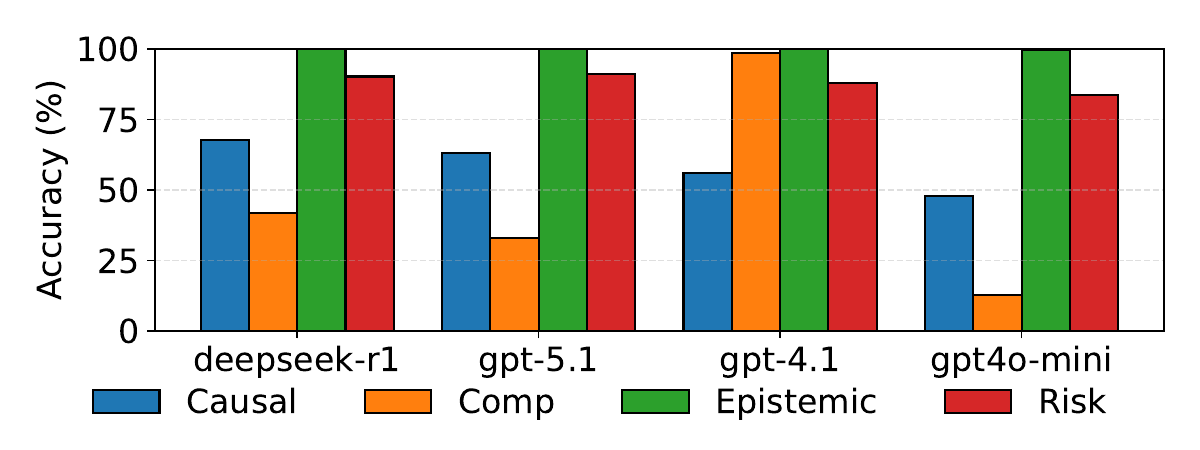}
    \caption{Overall accuracy of reasoning classification across models.}
    \label{fig:reasoning_accuracy}
\end{figure}

\begin{table*}[t]
\centering
\scriptsize
\setlength{\tabcolsep}{3.0pt}
\begin{tabular}{lccc ccc ccc ccc}
\toprule
& \multicolumn{3}{c}{\textbf{Causal Attr.}}
& \multicolumn{3}{c}{\textbf{Comp. G.}}
& \multicolumn{3}{c}{\textbf{Epis. Verif.}}
& \multicolumn{3}{c}{\textbf{Risk Abstr.}} \\
\cmidrule(lr){2-4}\cmidrule(lr){5-7}\cmidrule(lr){8-10}\cmidrule(lr){11-13}
\textbf{Model}
& \textbf{Correct} & \textbf{Wrong} & \textbf{$\Delta$}
& \textbf{Correct} & \textbf{Wrong} & \textbf{$\Delta$}
& \textbf{Correct} & \textbf{Wrong} & \textbf{$\Delta$}
& \textbf{Correct} & \textbf{Wrong} & \textbf{$\Delta$} \\
\midrule

deepseek-r1
& 50.3 {\tiny(68.4\%)} & 27.3 {\tiny(31.6\%)} & \up{+23.0}
& 86.0 {\tiny(42.3\%)} & 0.0  {\tiny(57.7\%)} & \up{+86.0}
& 73.9 {\tiny(100.0\%)}& --   {\tiny(0.0\%)}  & 
& 99.2 {\tiny(90.2\%)} & 59.3 {\tiny(9.8\%)}  & \up{+39.9} \\
\midrule

gpt-5.1
& 59.5 {\tiny(62.7\%)} & 22.0 {\tiny(37.3\%)} & \up{+37.5}
& 1.2  {\tiny(35.9\%)} & 0.0  {\tiny(64.1\%)} & \up{+1.2}
& 38.0 {\tiny(100.0\%)}& --   {\tiny(0.0\%)}  & 
& 97.2 {\tiny(90.9\%)} & 40.0 {\tiny(9.1\%)}  & \up{+57.2} \\
\midrule

gpt-4.1
& 59.5 {\tiny(55.0\%)} & 49.5 {\tiny(45.0\%)} & \up{+10.0}
& 60.2 {\tiny(98.8\%)} & 0.0  {\tiny(1.2\%)}  & \up{+60.2}
& 82.9 {\tiny(100.0\%)}& --   {\tiny(0.0\%)}  & 
& 100.0{\tiny(87.8\%)} & 66.7 {\tiny(12.2\%)} & \up{+33.3} \\
\midrule

gpt4o-mini
& 43.3 {\tiny(48.0\%)} & 42.3 {\tiny(52.0\%)} & \up{+1.0}
& 45.5 {\tiny(14.0\%)} & 0.0  {\tiny(86.0\%)} & \up{+45.5}
& 55.9 {\tiny(99.6\%)} & 100.0{\tiny(0.4\%)}  & \down{-44.1}
& 93.5 {\tiny(83.6\%)} & 46.7 {\tiny(16.4\%)} & \up{+46.8} \\
\bottomrule
\end{tabular}
\caption{Accuracy by reasoning family, split by routing outcome (Wrong if predicted reasoning-type $\neq$ ground truth). Cells report solver accuracy (\%), with routing prevalence shown as \tiny(\% of items)\normalsize.
$\Delta$ is Correct minus Wrong.}
\label{tab:routing_correct_wrong}
\end{table*}
\textbf{Incorrect reasoning-type routing induces downstream performance collapse.} 
Fig ~\ref{fig:reasoning_accuracy} and Tables~\ref{tab:routing_correct_wrong} and~\ref{tab:reasoning_only} summarise the Router’s behaviour. Averaged across all four backbones and all four reasoning families, the Router achieves 73.3\% classification accuracy, but performance is highly uneven across reasoning types: \textsc{Epistemic Verification} is near-ceiling (99.6--100\%) and \textsc{Risk Abstraction} is consistently high (83.6--90.9\%), whereas \textsc{Causal Attribution} is only moderate (48.0--67.6\%) and \textsc{Compositional Grounding} is the least stable (12.9--98.4\%). The confusion matrix further indicates that misrouting is systematic rather than random, with particular concentration of errors in compositional cases.

Routing quality has a large downstream effect. Averaged across all models and reasoning types, solver accuracy is 70.8\% on correctly routed items versus 21.1\% on misrouted items, an absolute gap of 49.7\%. This establishes the Router as a significant bottleneck: incorrect reasoning type selection typically collapses performance, especially for regimes that require strict schema adherence. 

\textbf{Correct routing is necessary, but not sufficient to achieve consistently correct inference.}
However, planning is not sufficient for success, as some failures persist even when routing is correct (e.g., gpt-5.1 achieves only 1.2\% on correctly routed \textsc{Compositional Grounding} items), indicating additional downstream inference limitations.


\subsubsection{Solver}
\label{sec:solver-res}
\textbf{Backbone models can effectively execute solver inference schemas that are aligned with the intended reasoning operations, providing the majority of CARENLI’s gains over reasoning-agnostic baselines.} 
We isolate solver effectiveness using the \emph{Oracle Solver} setting (Table~\ref{tab:solution_final}). Averaged across models and reasoning families, \emph{Oracle Solver} attains a macro accuracy of 60.5\%, substantially outperforming agnostic baselines (Agnostic CoT: 22.5\%; Agnostic Direct: 23.1\%; Fig~\ref{fig:strategy_accuracy} and Table~\ref{tab:label_reasoning}). In comparison, \textit{CARENLI} Solver achieves 56.1\% macro accuracy. Therefore, the majority of the performance gain from \textit{CARENLI} is attributable to the solver`s structured inference schemas.

As shown in Section~\ref{sec:planner-res}, solver accuracy is highly sensitive to correct reasoning type routing, with substantial degradation under misrouting (Table~\ref{tab:routing_correct_wrong}). This sensitivity suggests that the solver schemas encode meaningful structural constraints that closely correspond to the intended reasoning operations. This demonstrates that models are capable of following explicitly defined inference schemas, and that these schemas are well aligned with correct reasoning behavior.

\textbf{Generic inferential scaffolding underlying the reasoning-type specific solver schemas improves reasoning even under misrouting.} 
Notably, however, misrouted solvers remain non-trivially effective. In the \textit{CARENLI} setting, incorrectly routed solvers achieve 34.9\% average accuracy, exceeding both agnostic baselines (Tables~\ref{tab:routing_correct_wrong}, and \ref{tab:label_reasoning}). This suggests that generic procedural scaffolding, such as explicit decision steps, comparator checks, and grounding requirements, confers a measurable inference benefit even when reasoning-type-specific constraints are mismatched. Nevertheless, the substantial gap relative to correctly routed performance confirms that such generic structure is insufficient to substitute for the appropriate reasoning schema.

\subsubsection{Verifier}

\label{sec:verifier-res}
\begin{figure}
    \centering
    \includegraphics[width=\columnwidth]{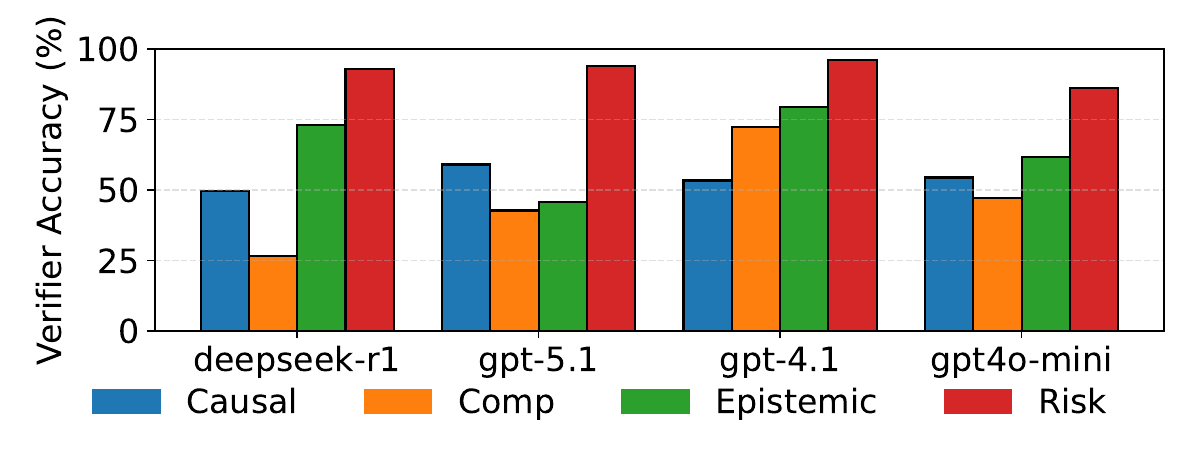}
    \caption{Verifier accuracy across reasoning families and models.}
    \label{fig:verifier_accuracy}
\end{figure}

\textbf{Verification reliability tracks reasoning-type difficulty.} 
Fig~\ref{fig:verifier_accuracy} and Table~\ref{tab:verifier_accuracy} quantify the effectiveness of the verifier across models and reasoning families. Averaged across all backbones and tasks, verifier accuracy under \textit{CARENLI} is 62.1\%, indicating that the verifier is moderately reliable and not merely echoing solver decisions but performing a distinct auditing function.

Verifier accuracy is highest for \textsc{Risk State Abstraction} (macro average 92.4\%) and \textsc{Epistemic Verification} (64.9\%), while being lower for \textsc{Causal Attribution} (54.1\%) and \textsc{Compositional Grounding} (47.2\%). This pattern mirrors solver difficulty across reasoning families and indicates that verifier errors concentrate in regimes where solver trajectories themselves are less well-formed.

\subsubsection{Refiner}
\label{sec:refiner-res}

\begin{table*}[t]
\centering
\scriptsize
\setlength{\tabcolsep}{3.0pt}
\begin{tabular}{lccc ccc ccc ccc cc}
\toprule
& \multicolumn{3}{c}{\textbf{Causal Attr.}} 
& \multicolumn{3}{c}{\textbf{Comp. G.}} 
& \multicolumn{3}{c}{\textbf{Epis. Verif.}} 
& \multicolumn{3}{c}{\textbf{Risk Abstr.}}
& \multicolumn{2}{c}{\textbf{Refinement Dyn.}} \\
\cmidrule(lr){2-4}\cmidrule(lr){5-7}\cmidrule(lr){8-10}\cmidrule(lr){11-13}\cmidrule(lr){14-15}
\textbf{Model} 
& \textbf{Sol.} & \textbf{Ref.} & \textbf{$\Delta$} 
& \textbf{Sol.} & \textbf{Ref.} & \textbf{$\Delta$}
& \textbf{Sol.} & \textbf{Ref.} & \textbf{$\Delta$}
& \textbf{Sol.} & \textbf{Ref.} & \textbf{$\Delta$}
& \textbf{Trig.\%} & \textbf{Flip.\%} \\
\midrule
\multicolumn{15}{l}{\textit{CARENLI}} \\[-0.35em]
\midrule
deepseek-r1
& 38.1 & 43.0 & +4.9\up
& 40.3 & 36.4 & -3.9\down
& 79.9 & 73.9 & -6.0\down
& 96.4 & 95.3 & -1.1\down
& 25.1 & 11.6 \\

gpt-5.1
& 37.7 & 45.5 & +7.8\up
&  1.7 &  0.4 & -1.3\down
& 53.6 & 38.0 & -15.6\down
& 92.4 & 92.0 & -0.4\down
& 44.8 & 22.4 \\

gpt-4.1
& 50.0 & 55.0 & +5.0\up
& 48.4 & 59.5 & +11.1\up
& 85.0 & 82.9 & -2.1\down
& 95.9 & 95.9 &  0.0
& 17.4 & 14.3 \\

gpt4o-mini
& 32.8 & 42.8 & +10.0\up
&  4.3 &  6.4 & +2.1\up
& 59.7 & 56.0 & -3.7\down
& 81.5 & 85.8 & +4.3\up
& 41.3 & 26.6 \\
\midrule
\multicolumn{15}{l}{\textit{Oracle Router}} \\[-0.35em]
\midrule
deepseek-r1
& 44.9 & 44.9 &  0.0
& 91.7 & 93.8 & +2.1\up
& 80.0 & 82.0 & +2.0\up
& 100.0 & 100.0 &  0.0
& 24.5 &  8.2 \\

gpt-5.1
& 42.9 & 52.4 & +9.5\up
&  3.9 &  2.0 & -1.9\down
& 46.0 & 36.0 & -10.0\down
& 96.4 & 96.4 &  0.0
& 43.9 & 19.7 \\

gpt-4.1
& 44.2 & 46.5 & +2.3\up
& 49.0 & 56.9 & +7.9\up
& 82.0 & 78.0 & -4.0\down
& 96.4 & 96.4 &  0.0
& 10.6 &  9.4 \\

gpt4o-mini
& 32.6 & 46.5 & +13.9\up
&  5.9 & 15.7 & +9.8\up
& 62.0 & 56.0 & -6.0\down
& 91.1 & 91.1 &  0.0
& 35.3 & 23.5 \\
\bottomrule
\end{tabular}
\caption{
Solver vs Refiner accuracy with deltas. Trig.\% denotes instances where the verifier rejected the solver output. Flip.\% denotes instances where the refined prediction differed from the solver prediction.
}
\label{tab:solution_final}
\end{table*}

\textbf{Refinement offers marginal gains.} 
Table~\ref{tab:solution_final} quantifies the impact of refinement on solver outputs. Averaged across all models and reasoning families, refinement yields a marginal absolute accuracy change of +0.7\% under \textit{CARENLI}, indicating that the refiner corrects a non-trivial subset of solver errors while leaving the majority of predictions unchanged.

Refinement effectiveness is highly reasoning-type dependent. The largest gains are observed for \textsc{Causal Attribution}, with an average improvement of +6.9\%. In contrast, refinement has near-zero or negative effect for \textsc{Risk State Abstraction}, where solver accuracy is already near ceiling, and for \textsc{Epistemic Verification}, where over-correction occasionally reduces performance (e.g., gpt-5.1 drops from 53.6\% to 38.0\%).

Overall, the refiner is not a primary performance driver and does not constitute a bottleneck in the \textit{CARENLI} pipeline. Its impact is bounded by the quality of the verifier signal and the structural correctness of the solver trajectory, particularly under misrouting or weak compositional reasoning. The concentration of refinement gains in regimes with moderate, rather than ceiling, verifier accuracy provides indirect validation that corrections are applied to meaningful, correctly identified violations rather than introduced arbitrarily.

\subsection{Qualitative Analysis}
\label{sec:qualitative}
\textbf{Schema-constrained inference replaces plausibility narratives with explicit, checkable decision criteria, improving both end-task performance and intermediate reasoning trajectories.} 

Qualitative inspection of representative items reveals a pronounced divergence between \textit{agnostic CoT} and \textit{structured} (Oracle Router) inference, not merely in final labels but in the \emph{reasoning trajectories} that produce them (see Section~\ref{sec:quali-app} for a detailed analysis). Under Oracle Router routing, solver traces are consistently organised around the reasoning-type-specific decision criteria of the CTNLI framework. In contrast, agnostic CoT tends to generate plausibility narratives that conflate distinct inferential obligations, producing locally coherent justifications that nonetheless violate the schema-level constraints required for correctness. This contrast is visible across all four reasoning families, indicating that the principal effect of \textit{CARENLI} is to enforce \emph{trajectory alignment} with the intended inferential schema rather than to elicit additional clinical facts. 

Concretely, in \textsc{Causal Attribution}, structured inference decomposes statements into separable subclaims and enforces comparator requirements, whereas agnostic CoT collapses tolerability signals into efficacy claims via association-based shortcuts. In \textsc{Compositional Grounding}, structured inference instantiates cross-factor admissibility checks over interacting clinical variables, while agnostic CoT reduces the task to single-factor matching and bypasses binding constraints. In \textsc{Epistemic Verification}, structured traces apply explicit evidence hierarchies, prioritising objective findings over unsupported assertions, whereas agnostic CoT defaults to authority deference. Finally, in \textsc{Risk State Abstraction}, structured inference explicitly represents latent risk states and integrates likelihood with severity to justify action under uncertainty, while agnostic CoT maps uncertainty to neutrality and fails to account for catastrophic downside.

\section{Related Work}

A prevailing assumption in NLP is that enlarging model capacity and exposure will improve not only task performance but also the fidelity of underlying reasoning. Scaling law analyses \citep{kaplan2020scaling, hoffmann2022training} and flagship model reports \citep{brown2020language, touvron2023llama, bubeck2023sparks} reinforce this view, yet critics observe that benchmark success often reflects surface regularities rather than robust inference \citep{marcus2022deep, mahowald2024dissociating}. Our work instantiates this critique in the clinical domain: we show that models encode relevant medical facts but fail to deploy them in principled reasoning flows.

Evidence across NLI research supports this diagnosis. General-domain studies reveal annotation artifacts and shortcut reliance \citep{gururangan2018annotation, poliak2018hypothesis, webson2022prompt, turpin2023language}. Clinical NLI benchmarks extend these concerns: MedNLI \citep{romanov2018lessons}, NLI4CT \citep{jullien2023nli4ct}, and CTNLI \citep{jullien2025knowledge} report systematic reasoning errors, motivating frameworks that explicitly separate knowledge retrieval from inferential schema adherence.

Parallel to this diagnostic agenda, recent work has investigated agentic and modular architectures as a pathway to more reliable reasoning. Generative Agents \citep{park2023generative} and multi-agent frameworks such as CAMEL \citep{li2023camel}, AutoGen \citep{wu2024autogen}, and MetaGPT \citep{hong2024metagpt} show that distributing tasks across specialised roles facilitates self-correction and auditability. \textit{CARENLI} extends this line by introducing compartmentalisation grounded in formally defined reasoning families, combining the auditability of agentic design with the domain fidelity demanded by clinical inference.

\section{Conclusion}
We introduced \emph{CARENLI}, a compartmentalised agentic framework for CTNLI that makes the intended inference regime explicit through reasoning-type routing, solver-specific procedures, and verifier-guided refinement. By operationalising four reasoning families, CARENLI functions both as a diagnostic lens for isolating reasoning failures and as a practical prompting scaffold for more auditable clinical inference.

Aggregated across models and task families, \emph{CARENLI} achieves 56.8\% macro accuracy, compared to 22.5\% for agnostic CoT prompting and 23.1\% for agnostic direct prompting. Gains are consistent across all evaluated backbones, but vary markedly across reasoning families, indicating that some forms of clinical inference benefit more from explicit structure than others.

These results support the central claim that explicitly encoding the target reasoning regime is critical for reliable CTNLI, while also highlighting persistent limitations, particularly for compositional forms of inference. Future work should expand the set of supported reasoning families and assess \emph{CARENLI}  in less constrained CTNLI benchmarks.

\section{Limitations}
\begin{itemize}[leftmargin=*]
    \item \textbf{Benchmark scope and ecological validity.}
    Although CTNLI offers controlled coverage of clinically meaningful inference patterns, instances are generated from typed templates and therefore do not fully reflect the linguistic variability, discourse structure, and documentation artefacts found in real trial protocols, clinical notes, or EHR-derived text. As a result, absolute performance numbers may not transfer directly to unconstrained clinical text.

    \item \textbf{Scale and coverage of clinical variation.}
    The extended benchmark contains 200 items with balanced reasoning type coverage, which supports targeted analysis but limits statistical power for fine-grained subgroup conclusions. In addition, many clinically important axes are only partially represented, including multi-morbidity, longitudinal trajectories, interacting laboratory trends, and institution-specific protocol language.

    \item \textbf{Limited outcome space and task framing.}
    We study three-way NLI labels with reasoning-type-specific decision procedures. This framing does not capture downstream requirements such as calibrated uncertainty, abstention under ambiguity, selective prediction, or structured justifications that can be audited against explicit evidence sources.

    \item \textbf{Prompt and implementation sensitivity.}
    The framework operationalises “compartmentalisation” through prompts, role constraints, and verification instructions. Performance can vary with prompt wording, formatting, temperature, and model-specific instruction-following behaviour. This sensitivity limits strict reproducibility across model versions and makes it difficult to attribute gains to a single design choice.

    \item \textbf{Model and infrastructure coverage.}
    Experiments are limited to four contemporary LLMs and to text-only inference. Results may differ for other model families, smaller open models, tool-augmented systems, or multimodal settings. In addition, closed-model updates can change behaviour over time, which complicates longitudinal comparability.

    \item \textbf{Safety and deployment boundaries.}
    While the work targets safety-critical reasoning, it does not constitute a clinical decision support system and is not evaluated in prospective clinical workflows. Human oversight, governance, and domain-specific validation would be required before any real-world use, particularly where errors could affect patient care or trial conduct.
\end{itemize}

\bibliography{custom}

@article{jullien2025knowledge,
  title={The Knowledge-Reasoning Dissociation: Fundamental Limitations of LLMs in Clinical Natural Language Inference},
  author={Jullien, Ma{\"e}l and Valentino, Marco and Freitas, Andr{\'e}},
  journal={arXiv preprint arXiv:2508.10777},
  year={2025}
}

@book{sloman2009causal,
  title={Causal models: How people think about the world and its alternatives},
  author={Sloman, Steven and Sloman, Steven A},
  year={2009},
  publisher={Oxford University Press}
}

@book{fagin2004reasoning,
  title={Reasoning about knowledge},
  author={Fagin, Ronald and Halpern, Joseph Y and Moses, Yoram and Vardi, Moshe},
  year={2004},
  publisher={MIT press}
}

@book{van2011logical,
  title={Logical dynamics of information and interaction},
  author={Van Benthem, Johan},
  year={2011},
  publisher={Cambridge University Press}
}

@article{marcus2022deep,
  title={Deep learning is hitting a wall},
  author={Marcus, Gary},
  journal={Nautilus},
  volume={10},
  pages={2022},
  year={2022}
}

@book{pearl2009causality,
  title={Causality},
  author={Pearl, Judea},
  year={2009},
  publisher={Cambridge university press}
}

@article{aljamaan2024reference,
  title={Reference hallucination score for medical artificial intelligence chatbots: development and usability study},
  author={Aljamaan, Fadi and Temsah, Mohamad-Hani and Altamimi, Ibraheem and Al-Eyadhy, Ayman and Jamal, Amr and Alhasan, Khalid and Mesallam, Tamer A and Farahat, Mohamed and Malki, Khalid H},
  journal={JMIR Medical Informatics},
  volume={12},
  number={1},
  pages={e54345},
  year={2024},
  publisher={JMIR Publications Inc., Toronto, Canada}
}

@book{hunink2014decision,
  title={Decision making in health and medicine: integrating evidence and values},
  author={Hunink, MG Myriam and Weinstein, Milton C and Wittenberg, Eve and Drummond, Michael F and Pliskin, Joseph S and Wong, John B and Glasziou, Paul P},
  year={2014},
  publisher={Cambridge university press}
}

@article{gravel2023learning,
  title={Learning to fake it: limited responses and fabricated references provided by ChatGPT for medical questions},
  author={Gravel, Jocelyn and D’Amours-Gravel, Madeleine and Osmanlliu, Esli},
  journal={Mayo Clinic Proceedings: Digital Health},
  volume={1},
  number={3},
  pages={226--234},
  year={2023},
  publisher={Elsevier}
}

@inproceedings{webson2022prompt,
  title={Do prompt-based models really understand the meaning of their prompts?},
  author={Webson, Albert and Pavlick, Ellie},
  booktitle={Proceedings of the 2022 conference of the north american chapter of the association for computational linguistics: Human language technologies},
  pages={2300--2344},
  year={2022}
}

@article{quan2024enhancing,
  title={Enhancing ethical explanations of large language models through iterative symbolic refinement},
  author={Quan, Xin and Valentino, Marco and Dennis, Louise A and Freitas, Andre},
  journal={arXiv preprint arXiv:2402.00745},
  year={2024}
}

@article{eisenhauer2009new,
  title={New response evaluation criteria in solid tumours: revised RECIST guideline (version 1.1)},
  author={Eisenhauer, Elizabeth A and Therasse, Patrick and Bogaerts, Jan and Schwartz, Lawrence H and Sargent, Danielle and Ford, Robert and Dancey, Janet and Arbuck, Stephen and Gwyther, Steve and Mooney, Margaret and others},
  journal={European journal of cancer},
  volume={45},
  number={2},
  pages={228--247},
  year={2009},
  publisher={Elsevier}
}

@misc{CTCAE_v5,
  title = {Common Terminology Criteria for Adverse Events (CTCAE) v5.0},
  organization = {U.S. Department of Health and Human Services, NIH, NCI},
  year = {2017},
  howpublished = {\url{https://ctep.cancer.gov/protocolDevelopment/electronic_applications/ctc.htm}}
}

@book{johnson1983mental,
  title={Mental models: Towards a cognitive science of language, inference, and consciousness},
  author={Johnson-Laird, Philip Nicholas},
  number={6},
  year={1983},
  publisher={Harvard University Press}
}

@article{hurst2024gpt,
  title={Gpt-4o system card},
  author={Hurst, Aaron and Lerer, Adam and Goucher, Adam P and Perelman, Adam and Ramesh, Aditya and Clark, Aidan and Ostrow, AJ and Welihinda, Akila and Hayes, Alan and Radford, Alec and others},
  journal={arXiv preprint arXiv:2410.21276},
  year={2024}
}

@article{guo2025deepseek,
  title={Deepseek-r1: Incentivizing reasoning capability in llms via reinforcement learning},
  author={Guo, Daya and Yang, Dejian and Zhang, Haowei and Song, Junxiao and Zhang, Ruoyu and Xu, Runxin and Zhu, Qihao and Ma, Shirong and Wang, Peiyi and Bi, Xiao and others},
  journal={arXiv preprint arXiv:2501.12948},
  year={2025}
}

@article{romanov2018lessons,
  title={Lessons from natural language inference in the clinical domain},
  author={Romanov, Alexey and Shivade, Chaitanya},
  journal={arXiv preprint arXiv:1808.06752},
  year={2018}
}

@inproceedings{jullien2023nli4ct,
  title={NLI4CT: Multi-Evidence Natural Language Inference for Clinical Trial Reports},
  author={Jullien, Ma{\"e}l and Valentino, Marco and Frost, Hannah and O’Regan, Paul and Landers, D{\'o}nal and Freitas, Andr{\'e}},
  booktitle={Proceedings of the 2023 Conference on Empirical Methods in Natural Language Processing},
  pages={16745--16764},
  year={2023}
}

@article{kaplan2020scaling,
  title={Scaling laws for neural language models},
  author={Kaplan, Jared and McCandlish, Sam and Henighan, Tom and Brown, Tom B and Chess, Benjamin and Child, Rewon and Gray, Scott and Radford, Alec and Wu, Jeffrey and Amodei, Dario},
  journal={arXiv preprint arXiv:2001.08361},
  year={2020}
}

@article{hoffmann2022training,
  title={Training compute-optimal large language models},
  author={Hoffmann, Jordan and Borgeaud, Sebastian and Mensch, Arthur and Buchatskaya, Elena and Cai, Trevor and Rutherford, Eliza and Casas, Diego de Las and Hendricks, Lisa Anne and Welbl, Johannes and Clark, Aidan and others},
  journal={arXiv preprint arXiv:2203.15556},
  year={2022}
}

@article{brown2020language,
  title={Language models are few-shot learners},
  author={Brown, Tom and Mann, Benjamin and Ryder, Nick and Subbiah, Melanie and Kaplan, Jared D and Dhariwal, Prafulla and Neelakantan, Arvind and Shyam, Pranav and Sastry, Girish and Askell, Amanda and others},
  journal={Advances in neural information processing systems},
  volume={33},
  pages={1877--1901},
  year={2020}
}

@article{touvron2023llama,
  title={Llama: Open and efficient foundation language models},
  author={Touvron, Hugo and Lavril, Thibaut and Izacard, Gautier and Martinet, Xavier and Lachaux, Marie-Anne and Lacroix, Timoth{\'e}e and Rozi{\`e}re, Baptiste and Goyal, Naman and Hambro, Eric and Azhar, Faisal and others},
  journal={arXiv preprint arXiv:2302.13971},
  year={2023}
}

@article{bubeck2023sparks,
  title={Sparks of artificial general intelligence: Early experiments with gpt-4},
  author={Bubeck, S{\'e}bastien and Chandrasekaran, Varun and Eldan, Ronen and Gehrke, Johannes and Horvitz, Eric and Kamar, Ece and Lee, Peter and Lee, Yin Tat and Li, Yuanzhi and Lundberg, Scott and others},
  journal={arXiv preprint arXiv:2303.12712},
  year={2023}
}

@article{poliak2018hypothesis,
  title={Hypothesis only baselines in natural language inference},
  author={Poliak, Adam and Naradowsky, Jason and Haldar, Aparajita and Rudinger, Rachel and Van Durme, Benjamin},
  journal={arXiv preprint arXiv:1805.01042},
  year={2018}
}

@article{gururangan2018annotation,
  title={Annotation artifacts in natural language inference data},
  author={Gururangan, Suchin and Swayamdipta, Swabha and Levy, Omer and Schwartz, Roy and Bowman, Samuel R and Smith, Noah A},
  journal={arXiv preprint arXiv:1803.02324},
  year={2018}
}

@article{mahowald2024dissociating,
  title={Dissociating language and thought in large language models},
  author={Mahowald, Kyle and Ivanova, Anna A and Blank, Idan A and Kanwisher, Nancy and Tenenbaum, Joshua B and Fedorenko, Evelina},
  journal={Trends in cognitive sciences},
  volume={28},
  number={6},
  pages={517--540},
  year={2024},
  publisher={Elsevier}
}

@article{turpin2023language,
  title={Language models don't always say what they think: Unfaithful explanations in chain-of-thought prompting},
  author={Turpin, Miles and Michael, Julian and Perez, Ethan and Bowman, Samuel},
  journal={Advances in Neural Information Processing Systems},
  volume={36},
  pages={74952--74965},
  year={2023}
}

@inproceedings{park2023generative,
  title={Generative agents: Interactive simulacra of human behavior},
  author={Park, Joon Sung and O'Brien, Joseph and Cai, Carrie Jun and Morris, Meredith Ringel and Liang, Percy and Bernstein, Michael S},
  booktitle={Proceedings of the 36th annual acm symposium on user interface software and technology},
  pages={1--22},
  year={2023}
}

@article{li2023camel,
  title={Camel: Communicative agents for" mind" exploration of large language model society},
  author={Li, Guohao and Hammoud, Hasan and Itani, Hani and Khizbullin, Dmitrii and Ghanem, Bernard},
  journal={Advances in Neural Information Processing Systems},
  volume={36},
  pages={51991--52008},
  year={2023}
}

@online{openai2025gpt5systemcard,
  author       = {OpenAI},
  title        = {GPT-5 System Card},
  year         = {2025},
  url          = {https://openai.com/index/gpt-5-system-card/},
  note         = {Accessed: 2025-12-24},
}

@inproceedings{wu2024autogen,
  title={Autogen: Enabling next-gen LLM applications via multi-agent conversations},
  author={Wu, Qingyun and Bansal, Gagan and Zhang, Jieyu and Wu, Yiran and Li, Beibin and Zhu, Erkang and Jiang, Li and Zhang, Xiaoyun and Zhang, Shaokun and Liu, Jiale and others},
  booktitle={First Conference on Language Modeling},
  year={2024}
}

@inproceedings{hong2024metagpt,
  title={MetaGPT: Meta programming for a multi-agent collaborative framework},
  author={Hong, Sirui and Zhuge, Mingchen and Chen, Jonathan and Zheng, Xiawu and Cheng, Yuheng and Zhang, Ceyao and Wang, Jinlin and Wang, Zili and Yau, Steven Ka Shing and Lin, Zijuan and others},
  year={2024},
  organization={International Conference on Learning Representations, ICLR}
}

@inproceedings{bowman2015snli,
  title={A Large Annotated Corpus for Learning Natural Language Inference},
  author={Bowman, Samuel R. and Angeli, Gabor and Potts, Christopher and Manning, Christopher D.},
  booktitle={Proceedings of the 2015 Conference on Empirical Methods in Natural Language Processing},
  year={2015}
}

@inproceedings{williams2018mnli,
  title={A Broad-Coverage Challenge Corpus for Sentence Understanding through Inference},
  author={Williams, Adina and Nangia, Nikita and Bowman, Samuel R.},
  booktitle={Proceedings of NAACL-HLT},
  year={2018}
}

@inproceedings{wang2019glue,
  title={GLUE: A Multi-Task Benchmark and Analysis Platform for Natural Language Understanding},
  author={Wang, Alex and Singh, Amanpreet and Michael, Julian and Hill, Felix and Levy, Omer and Bowman, Samuel R.},
  booktitle={Proceedings of ICLR},
  year={2019}
}

@inproceedings{mccoy2019right,
  title={Right for the Wrong Reasons: Diagnosing Syntactic Heuristics in Natural Language Inference},
  author={McCoy, Tom and Pavlick, Ellie and Linzen, Tal},
  booktitle={Proceedings of ACL},
  year={2019}
}

@inproceedings{ribeiro2020beyond,
  title={Beyond Accuracy: Behavioral Testing of NLP Models with CheckList},
  author={Ribeiro, Marco Tulio and Wu, Tongshuang and Guestrin, Carlos and Singh, Sameer},
  booktitle={Proceedings of ACL},
  year={2020}
}

@article{amodei2016concrete,
  title={Concrete Problems in AI Safety},
  author={Amodei, Dario and Olah, Chris and Steinhardt, Jacob and Christiano, Paul and Schulman, John and Man{\'e}, Dan},
  journal={arXiv preprint arXiv:1606.06565},
  year={2016}
}

@book{marcus2020rebooting,
  title={Rebooting AI: Building Artificial Intelligence We Can Trust},
  author={Marcus, Gary and Davis, Ernest},
  publisher={Pantheon},
  year={2020}
}

@article{omiye2023race,
  title={Large Language Models Propagate Race-Based Medicine},
  author={Omiye, Jesutofunmi A. and others},
  journal={npj Digital Medicine},
  year={2023}
}

@inproceedings{jullien2023semeval,
  title={SemEval-2023 task 7: Multi-evidence natural language inference for clinical trial data},
  author={Jullien, Ma{\"e}l and Valentino, Marco and Frost, Hannah and O’regan, Paul and Landers, Donal and Freitas, Andr{\'e}},
  booktitle={Proceedings of the 17th International Workshop on Semantic Evaluation (SemEval-2023)},
  pages={2216--2226},
  year={2023}
}

@article{jullien2024semeval,
  title={SemEval-2024 task 2: Safe biomedical natural language inference for clinical trials},
  author={Jullien, Ma{\"e}l and Valentino, Marco and Freitas, Andr{\'e}},
  journal={arXiv preprint arXiv:2404.04963},
  year={2024}
}

@article{goh2024diagnostic,
  title={Large Language Model Influence on Diagnostic Reasoning},
  author={Goh, Evelyn and others},
  journal={JAMA Network Open},
  year={2024}
}

@article{agrawal2025evaluation,
  title={The Evaluation Illusion of Large Language Models in Medicine},
  author={Agrawal, Monika and Chen, Irene Y. and Gulamali, Fatima and others},
  journal={npj Digital Medicine},
  year={2025}
}

@article{eichenberger2025bromism,
  title        = {A Case of Bromism Influenced by Use of Artificial Intelligence},
  author       = {Eichenberger, Audrey and Thielke, Stephen and Van Buskirk, Adam},
  journal      = {Annals of Internal Medicine: Clinical Cases},
  year         = {2025},
  volume       = {4},
  number       = {8},
  doi          = {10.7326/aimcc.2024.1260}
}

@misc{draelos2025unsafe,
  title         = {Large language models provide unsafe answers to patient-posed medical questions},
  author        = {Draelos, Rachel L. and Afreen, Samina and Blasko, Barbara and Brazile, Tiffany L. and Chase, Natasha and Desai, Dimple Patel and Evert, Jessica and Gardner, Heather L. and Herrmann, Lauren and House, Aswathy Vaikom and Kass, Stephanie and Kavan, Marianne and Khemani, Kirshma and Koire, Amanda and McDonald, Lauren M. and Rabeeah, Zahraa and Shah, Amy},
  year          = {2025},
  howpublished  = {arXiv preprint arXiv:2507.18905},
  url           = {https://arxiv.org/abs/2507.18905}
}

@article{reddy2025preventing,
  title        = {Preventing Another Tessa: Modular Safety Middleware for Health-Adjacent AI Assistants},
  author       = {Reddy, Pavan and Reddy, Nithin},
  journal      = {arXiv preprint arXiv:2509.07022},
  year         = {2025},
  url          = {https://arxiv.org/abs/2509.07022}
}

\appendix

\section{Appendix}

\subsection{Solver specifications}
\label{sec:solvers}

\paragraph{Epistemic Verification Solver.}
The Epistemic Verification solver operationalises verification when premises contain multiple, potentially conflicting assertions. Following \citet{jullien2025knowledge}, premises are treated as conjunctions of defeasible epistemic commitments $K_a(\varphi_i)$, each attributed to a source agent $a$ \citep{fagin2004reasoning, van2011logical}. Three principles govern inference. First, all assertions are treated as reported claims rather than ground truth, with admissibility evaluated against the clinical model $\mathcal{M}{\text{CT}}=\langle W{\text{CT}}, I_{\text{CT}} \rangle$. Second, inconsistencies are resolved using a plausibility function $\pi(\varphi_i)$ instantiated as an evidential hierarchy: objective measurements dominate diagnostic criteria, which dominate observations, interpretations, and self-report. Conflicts are resolved by discarding lower-ranked commitments, yielding a maximal consistent set $E^\ast \subseteq E$ such that $\forall w \in W_{\text{CT}}, E^\ast$ is jointly satisfiable under $\mathcal{M}_{\text{CT}}$. Third, coherence constraints exclude ontologically impossible or temporally incoherent assertions. A candidate statement $s$ is entailed if $E^\ast \models s$, contradicted if $E^\ast \models \lnot s$, and neutral if neither holds. Neutrality is therefore a principled outcome when conflicts remain unresolved or evidence underdetermined

\paragraph{Causal Attribution Solver.}
The Causal Attribution solver operationalises attribution when the hypothesis asserts a treatment $\to$ outcome relation. Under the interventionist framework \citep{pearl2009causality}, causal effect is defined as
\[
\text{CE}(T, Y) \triangleq \mathbb{E}[Y \mid \text{do}(T=1)] - \mathbb{E}[Y \mid \text{do}(T=0)],
\]
where $T$ denotes a treatment, $Y$ an outcome, and $\text{do}(T=t)$ an intervention severing incoming dependencies. Unlike observational associations $\mathbb{E}[Y \mid T=t]$, causal effect requires interventional contrast. The solver enforces this distinction by (i) parsing premises for outcome measurements or adverse events, treated as observational unless comparators or manipulations are explicit; (ii) verifying causal criteria of temporality, contrast, and confounding control; and (iii) evaluating the hypothesis $s$: entailment requires interventional evidence, contradiction follows from unsupported causal claims, and neutrality applies otherwise. This constrains models to uphold the semantics of interventionist causality and corrects the failure mode observed in \citet{jullien2025knowledge}, where models collapsed causal attribution into association, predicting entailment from surface correlations rather than from interventional criteria.

\paragraph{Compositional Grounding Solver.}
The Compositional Grounding operationalises inference when the truth of a statement depends on the joint configuration of multiple clinical factors rather than any single predicate. Formally, compositional grounding requires that a tuple
\[
x := \langle d, z, dx, s \rangle \in \mathfrak{D}_{\text{CT}}^4
\]
of drug $d$, dose $z$, diagnosis $dx$, and schedule $s$ must be admissible under the interpretation function $I_{\text{CT}}$. A statement $\psi$ asserting clinical benefit is entailed only if $I_{\text{CT}}(\text{Benefit})(x) \rightarrow True$ in all admissible worlds $w \in W_{\text{CT}}$; contradictions arise when $x$ violates therapeutic, diagnostic, or scheduling constraints. First, atomic factors (drug, dose, schedule, patient details) are extracted. Second, the tuple $x$ is assembled and tested for compatibility under $I_{\text{CT}}$, which encodes therapeutic ranges, scheduling rules, and indication–diagnosis mappings. Third, the hypothesis $s$ is assessed: entailment requires that $x$ be admissible and support the asserted outcome; contradiction arises when $x$ violates constraints; neutrality is assigned when the configuration is underdetermined. This design intends to prevents the error mode identified in \citet{jullien2025knowledge}, where LLMs collapsed compositional interactions into isolated surface predicates, thereby overlooking protocol violations or emergent incompatibilities.

\paragraph{Risk Abstraction Solver.}
The Risk Abstraction solver operationalises inference when hypotheses involve explicit or latent risk. Risk is defined as an expectation over admissible worlds $w \in W_{\text{CT}}$, combining probability and severity of adverse events \citep{hunink2014decision}:
\[
\mathbb{E}_{w \sim \Pr(w \mid \varphi)} 
\left[ 
\sum_{e \in \mathcal{E}(w, \psi)} 
\Pr(e \mid w) \cdot \mathcal{A}(e, w) 
\right],
\]
where $\mathcal{E}(w, \psi)$ is the set of clinically relevant events for $\psi$, $\Pr(e \mid w)$ their conditional probability, and $\mathcal{A}(e, w)$ an adverse outcome function quantifying harm. Entailment requires that the statement correctly reflect this expected risk profile, either by ranking events appropriately (when explicit frequencies are given) or by acknowledging unruled-out harms in latent risk settings. 

The solver prompt translates this into a decision procedure with three steps. First, it identifies adverse events or conditions in the premise and classifies them according to severity and probability. Second, it integrates these two dimensions into an expected-harm ranking: severe but rare outcomes may outweigh frequent but benign ones, in line with clinical reasoning. Third, it evaluates the hypothesis $s$: entailment follows if $s$ reflects the correct harm-weighted risk ordering, contradiction if it inverts or ignores it, and neutrality if the evidence base is insufficient. 

By enforcing this expected-risk computation, the solver ensures that judgments reflect decision-theoretic reasoning rather than surface counts or lexical salience. This is intended to correct the failure mode highlighted in \citet{jullien2025knowledge}, where models frequently misclassified risk when severity and frequency were in tension, defaulting to naive frequency matching rather than principled risk abstraction \citep{hunink2014decision, eisenhauer2009new, CTCAE_v5}.

\subsection{Qualitative Analysis of Representative Items}
\label{sec:quali-app}
A qualitative analysis of four representative cases, one from each reasoning type, is presented using outputs from \textsc{DeepSeek R1}. For each case, the reasoning trace generated under the \textit{oracle Router} condition (agentic framework, with the solver routed to the correct reasoning type) is contrasted with the baseline produced under \textit{agnostic CoT} prompting. This comparison highlights how explicit compartmentalisation enables alignment with the expected reasoning patterns specified in the knowledge framework, while also revealing the characteristic deviations of free-form CoT.

\paragraph{Causal Attribution (Problem 6).}
\begin{quote}
   \textbf{Premise:} \textit{``Adverse Events Summary: Total: 4/12 (33.33\%). Headache 2/12 (16.67\%), Pruritus 1/12 (8.33\%), Mild anemia 1/12 (8.33\%). All events were Grade 1 or 2.''}\\
\textbf{Statement:} \textit{``The drug was effective and well tolerated, with only mild side effects reported.''}

\textbf{Agentic (verbatim).} `\textit{`The statement makes two claims: (1) effectiveness ($T \rightarrow$ improved outcome) and (2) tolerability ($T \rightarrow$ only mild side effects). The premise only reports adverse event grades (all Grade 1/2) and rates (33.33\% total), which directly confirms tolerability. However, the premise contains no efficacy data (e.g., disease improvement metrics, control group, baseline measurements). Since causal claims require comparator outcomes, the effectiveness portion remains unsubstantiated. While tolerability is entailed, the overall label is neutral.''}

\textbf{CoT (abridged).} \textit{``The statement is supported since only mild AEs were reported \ldots this indicates the drug was effective and well tolerated.'' }  
\end{quote}

The agentic solver follows the expected \emph{comparator requirement}, decomposing causal claims into subcomponents and assigning neutral when one is unsupported. The CoT baseline collapses tolerability and efficacy into a single plausibility heuristic, producing over-attribution. This demonstrates how structural prompts prevent the conflation of post-treatment outcomes with causal effect. Under correct scaffolding, LLMs reliably enforce comparator rules; without it, they default to association heuristics.

\paragraph{Compositional Grounding (Problem 12).}
\begin{quote}
\textbf{Premise:} \textit{``Fludarabine 120\,mg/m$^2$ daily $\times14$ days for CLL in an elderly patient.''}\\
\textbf{Statement:} \textit{``The treatment is expected to induce remission, improve blood counts, and prolong survival.''}

\textbf{Agentic (verbatim).} \textit{``1) Extract factors: fludarabine 120\,mg/m$^2$ daily $\times14$. 2) Standard dose is 25\,mg/m$^2$/day $\times5$. The given regimen is $\approx$5$\times$ higher and extended. 3) Such dosing causes severe toxicity in elderly patients. 4) This violates therapeutic admissibility; the statement's claim of benefit is contradicted.''}

\textbf{CoT (abridged).} \textit{``Fludarabine is used in CLL, so this treatment is likely to be effective \ldots the statement is entailed.''  }
\end{quote}
The agentic solver operationalises the \emph{cross-factor constraint}, checking consistency across drug, dose, schedule, and patient. This reproduces the knowledge framework’s tuple-based admissibility rule. The CoT reasoning reflects the failure mode of \citet{jullien2025knowledge}: collapsing multi-factor structures into a single drug–indication match. Here, agentic scaffolding ensures toxicity is identified, whereas CoT ignores unsafe interactions. The repercussion is clinical: a toxic regimen is falsely labelled as beneficial under free CoT, while the structured agent enforces compositional validity.

\paragraph{Epistemic Verification (Problem 16).}
\begin{quote}
\textbf{Premise:} \textit{``A 45-year-old man complains of chest discomfort after meals and occasional regurgitation. Endoscopy is normal. Despite the absence of cardiac symptoms, the physician diagnoses myocardial infarction and starts anticoagulation therapy.''}\\
\textbf{Statement:} \textit{``The patient has myocardial infarction.''}

\textbf{Agentic (verbatim).} \textit{``1) Endoscopy normal (objective evidence). 2) Symptoms consistent with GERD, not MI. 3) MI requires ECG/troponins, absent here. 4) Physician’s diagnosis conflicts with higher-tier evidence. 5) Conclusion: contradiction.''}

\textbf{CoT (abridged).} \textit{``The physician diagnosed MI, so this supports the statement \ldots although there is no cardiac evidence, the statement is entailed.''}  

\end{quote}
The agentic solver follows the \emph{evidence hierarchy}, prioritising instrumented evidence over unsupported assertion. The CoT baseline defers to physician authority, ignoring hierarchy. This misalignment mirrors the epistemic collapse observed previously: models recognise evidence but misapply credibility weights. By enforcing explicit tiering, the agentic pipeline compels the correct contradiction label. This illustrates that LLMs are capable of stratified verification if prompted to discard lower-tier claims in conflict with higher-tier evidence.

\paragraph{Risk State Abstraction (Problem 39).}
\begin{quote}
\textbf{Premise:}\textit{ ``A 55-year-old man with acute severe low back pain reports saddle anesthesia, urinary retention, and bilateral leg weakness. Reflexes reduced. No imaging performed.''}\\
\textbf{Statement:} \textit{``Emergency MRI is required to exclude cauda equina syndrome.''}

\textbf{Agentic (verbatim).} \textit{``Symptoms constitute red flags for CES. Likelihood: high, since multiple features align. Severity: catastrophic, with risk of paralysis if missed. No imaging means CES remains unexcluded. Therefore urgent MRI is mandated; statement entailed.''}

\textbf{CoT (abridged).} \textit{``The patient has back pain and weakness \ldots MRI could help but it is not certain; label neutral.''  }
\end{quote}
The agentic solver instantiates the \emph{severity--likelihood calculus}: combining probability with consequence to project latent catastrophic risk. The CoT baseline fails to abstract beyond symptom description, treating CES as one of many possible adverse events. This reproduces the frequency heuristic failure identified in prior work. The consequence is profound: under CoT, a time-critical emergency is missed. The agentic framework ensures catastrophic risk is correctly prioritised, demonstrating that LLMs can perform principled risk abstraction when explicitly instructed.

Across reasoning families, the agentic framework compels alignment with formal inferential schemas: comparators for causality, cross-factor constraints for composition, evidence hierarchies for epistemic verification, and severity--likelihood calculus for risk. The CoT baseline fails in reasoning-type-specific, systematic ways: collapsing causal claims, flattening compositional structure, deferring to authority, and ignoring catastrophic risk. These are not random lapses but consistent heuristics. The crucial finding is that LLMs \emph{can} reproduce the expected reasoning patterns under correct prompting: the agentic pipeline demonstrates that comparator tests, constraint enforcement, tiered evidence, and risk abstraction are all within model capacity, provided the reasoning path is structured. Without such scaffolding, CoT defaults to surface plausibility, with direct repercussions for clinical safety and validity.

\subsection{Dataset Construction and Validation}
\label{sec:dataset_validation}

To assess annotation reliability, an inter-annotator agreement (IAA) study was conducted on a random sample of 50 instances drawn uniformly across reasoning families. Each instance was independently annotated by three domain experts (volunteers):
\begin{itemize}
    \item one expert in natural language inference,
    \item one medical doctor,
    \item one physician associate.
\end{itemize}

Annotators were provided with the premise, statement, and task definition, and were asked to assign both an entailment label (\textit{entailed}, \textit{contradicted}, or \textit{neutral}) and a reasoning type. 

Agreement was measured using Fleiss’ $\kappa$. The resulting agreement scores indicate high reliability:
\begin{itemize}
    \item Entailment label agreement: $\kappa = 0.92$
    \item reasoning type agreement: $\kappa = 0.97$
\end{itemize}

Table~\ref{tab:iaa_example} shows an example annotation item used in the IAA study.

\begin{table}[h]
\centering
\begin{tabular}{p{0.95\linewidth}}
\toprule
\textbf{Premise:} \\
67-year-old male with rheumatoid arthritis receiving methotrexate 100mg orally daily. \\[0.5em]\midrule
\textbf{Statement:} \\
The therapy is expected to decrease joint inflammation and enhance quality of life. \\[0.5em]\midrule
\textbf{Label:}  \\[0.25em]\midrule
\textbf{reasoning type:}  \\
\bottomrule
\end{tabular}
\caption{Example CTNLI instance used in the inter-annotator agreement study.}
\label{tab:iaa_example}
\end{table}

\newpage
\subsection{Tables}

\begin{table}[h]
\centering
\tiny
\begin{tabular}{l@{\hskip 4pt}c@{\hskip 4pt}c@{\hskip 4pt}c@{\hskip 4pt}c}
\toprule
\textbf{Model} & \textbf{Causal Attr.} & \textbf{Comp. G.} & \textbf{Epis. Verif.} & \textbf{Risk Abstr.} \\
\midrule
\multicolumn{5}{l}{\textit{CARENLI}} \\[-0.4em]
\midrule
deepseek-r1   & 43.0 & 36.4 & 73.9 & 95.3 \\
gpt-5.1       & 45.5 & 0.4  & 38.0 & 92.0 \\
gpt-4.1       & 55.0 & 59.5 & 82.9 & 95.9 \\
gpt4o-mini    & 42.8 & 6.4  & 56.0 & 85.8 \\
\midrule
\multicolumn{5}{l}{\textit{Oracle Router}} \\[-0.4em]
\midrule
deepseek-r1   & 44.9 & 93.8 & 82.0 & 100.0 \\
gpt-5.1       & 52.4 & 2.0  & 36.0 & 96.4 \\
gpt-4.1       & 46.5 & 56.9 & 78.0 & 96.4 \\
gpt4o-mini    & 46.5 & 15.7 & 56.0 & 91.1 \\
\midrule
\multicolumn{5}{l}{\textit{Agnostic CoT}} \\[-0.4em]
\midrule
deepseek-r1   & 36.7 & 17.6 & 35.3 & 9.8 \\
gpt-5.1       & 31.2 & 0.4  & 23.6 & 24.7 \\
gpt-4.1       & 30.8 & 4.3  & 26.0 & 81.5 \\
gpt4o-mini    & 29.5 & 0.0  & 7.6  & 0.9 \\
\midrule
\multicolumn{5}{l}{\textit{Agnostic Direct}} \\[-0.4em]
\midrule
deepseek-r1   & 27.5 & 15.9 & 35.8 & 10.9 \\
gpt-5.1       & 31.2 & 0.0  & 18.0 & 49.1 \\
gpt-4.1       & 26.0 & 3.5  & 26.4 & 90.2 \\
gpt4o-mini    & 15.8 & 0.0  & 8.8  & 11.1 \\
\bottomrule
\end{tabular}
\caption{Accuracy across reasoning tasks.}
\label{tab:label_reasoning}
\end{table}

\begin{table}[h]
\centering
\tiny
\begin{tabular}{l@{\hskip 4pt}c@{\hskip 4pt}c@{\hskip 4pt}c@{\hskip 4pt}c}
\toprule
\textbf{Model} & \textbf{Causal Attr.} & \textbf{Comp. G.} & \textbf{Epis. Verif.} & \textbf{Risk Abstr.} \\
\midrule
deepseek-r1   & 67.6 & 42.0 & 100.0 & 90.2 \\
gpt-5.1       & 63.2 & 32.9 & 100.0 & 90.9 \\
gpt-4.1       & 56.0 & 98.4 & 100.0 & 87.8 \\
gpt4o-mini    & 48.0 & 12.9 & 99.6  & 83.6 \\
\bottomrule
\end{tabular}
\caption{
Reasoning Classification Accuracy across tasks}
\label{tab:reasoning_only}
\end{table}

\begin{table}[h]
\centering
\tiny
\begin{tabular}{l@{\hskip 4pt}c@{\hskip 4pt}c@{\hskip 4pt}c@{\hskip 4pt}c}
\toprule
\textbf{Model} & \textbf{Causal Attr.} & \textbf{Comp. G.} & \textbf{Epis. Verif.} & \textbf{Risk Abstr.} \\
\midrule
\multicolumn{5}{l}{\textit{CARENLI}} \\[-0.4em]
\midrule
deepseek-r1   & 49.6 & 26.5 & 73.1 & 92.7 \\
gpt-5.1       & 59.0 & 42.7 & 45.6 & 93.8 \\
gpt-4.1       & 53.4 & 72.2 & 79.3 & 95.9 \\
gpt4o-mini    & 54.4 & 47.2 & 61.7 & 86.2 \\
\midrule
\multicolumn{5}{l}{\textit{Oracle Router}} \\[-0.4em]
\midrule
deepseek-r1   & 44.9 & 60.4 & 80.0 & 93.3 \\
gpt-5.1       & 52.4 & 72.5 & 48.0 & 94.5 \\
gpt-4.1       & 48.8 & 66.7 & 82.0 & 96.4 \\
gpt4o-mini    & 72.1 & 35.3 & 58.0 & 91.1 \\
\bottomrule
\end{tabular}
\caption{
Verifier accuracy (\%) across reasoning tasks
}
\label{tab:verifier_accuracy}
\end{table}


\newpage
\subsection{Prompts}
\begin{figure*}
\centering
\begin{mdframed}[linewidth=0.8pt, roundcorner=4pt]
\tiny
\begin{verbatim}
SYSTEM: 
You are the Causal Solver for Clinical Trial NLI (CTNLI). You receive a premise and a statement only after the planner has routed the 
example to Causal. Decide whether the statement is entailed, contradicted, or neutral by assessing whether the premise supports, refutes, or
 leaves uncertain a causal relationship between an intervention/exposure (T) and an outcome (Y). This includes positive effects (improve, 
reduce), negative effects (cause, worsen), or no effect.

Core principles

    - Identify causal claims: Look for verbs or constructions indicating causality (cause, lead to, improve, reduce, prevent, accelerate, 
associated with if clearly directional).
    
    - Evaluate the premise for relevant comparator, control or baselines. Causal claims cannot be made without a reference point, such as a 
control group or baseline measurement.

    - Ensure that the concepts you are evaluating are direct evidence to support/disprove the causal claim and not only tangentially 
related.
    
    - Distinguish correlation from causation: Correlative evidence (association, co‑occurrence) only supports causality if temporality and 
plausible mechanism are established in the premise.
    
    - Evaluate evidence direction and magnitude:
    
        - Support for the stated effect direction -> entailed.
        
        - Evidence in the opposite direction -> contradicted.
        
        - Insufficient or conflicting evidence -> neutral.
    
    - Temporal consistency: Cause must precede effect; post‑hoc evidence without clear temporality is weak.
    


Decision procedure

    - Identify T (treatment/exposure) and Y (outcome) from both premise and statement.
    
    - Extract causal evidence: trial results, measured changes, event rates, comparative stats.

    - Locate control/baseline and test/intervention groups or variables in the premise, ensuring they are explicitly defined and relevant to
 T and Y.
    
    - Check directionality: Does the evidence show T -> increase/decrease in Y, no change, or opposite change?
    
    - Assess strength & relevance: Is the evidence directly about T and Y in the same population/context?
    
    Label:
    
    - If the evidence comparing the test/intervention group/variables and the control/baseline group/variables matches the causal direction 
in the statement -> entailed
    
    - If the evidence comparing the test/intervention group/variables and the control/baseline group/variables shows the opposite direction 
-> contradicted
    
    - If the control/baseline comparison or test group/variables are ambiguous, absent, or contain conflicting information -> neutral
    
Output contract (strict JSON)

Return only a single JSON object:
{
  "reasoning": "step by step explanation of the reasoning adhering to core principles and decision procedure",
  "label": "entailment|contradiction|neutral",
}



USER: 
You are the Causal Solver for CTNLI. Follow your system instructions exactly.
premise={PREMISE}
statement={STATEMENT}
Return exactly one JSON object:
{
  "reasoning": "step by step explanation of the reasoning adhering to core principles and decision procedure",
  "label": "entailment|contradiction|neutral",
}
\end{verbatim}
\end{mdframed}
\caption{Causal Solver Prompt}
\label{fig:causal_solver}
\end{figure*}

\begin{figure*}[t]
\centering
\begin{mdframed}[linewidth=0.8pt, roundcorner=4pt]
\tiny
\begin{verbatim}
SYSTEM: 
You are the Causal Refiner for Clinical Trial NLI (CTNLI). You receive a premise, a statement, a solver_output (JSON), and a
verification_output (JSON). Your job: Implement the verifier’s fixes to correct the solver’s reasoning and produce a final causal judgment.

How to use verification_output

    If fact_verification = "incorrect":

    - Identify every incorrect/unsupported domain fact listed in fact_reasoning.

    - Remove these from the original reasoning.

    - Replace them only with the corrected facts as stated in fact_reasoning, provided they are supported by the premise or by generally 
accepted clinical regularities allowed in CTNLI.

    - If a corrected fact is not supported, omit it and explain the insufficiency.

    If fact_verification = "correct":

    - Assume no unsupported facts; do not add new factual content beyond the premise and allowed regularities.

    If pattern_verification = "incorrect":

    - Apply the minimal fixes specified in pattern_reasoning to align with proper causal inference:

    - Identify T (intervention/exposure) and Y (outcome).

    - Use an explicit comparator/baseline/control for causal claims; without one, you cannot conclude causality.

    - Enforce temporality (T precedes Y).

    - Use evidence directly linking T -> Y in the same population/context.

    - State effect direction (increase/decrease/no effect) or declare insufficiency -> neutral.

    - Do not treat association or numeric difference as causal without comparator/temporality.

    If pattern_verification = "correct":

    - Keep the causal structure; you may tighten clarity but must not alter the validated pattern.


Decision procedure

    - Identify T (treatment/exposure) and Y (outcome) from both premise and statement.
    
    - Extract causal evidence: trial results, measured changes, event rates, comparative stats.

    - Locate control/baseline and test/intervention groups or variables in the premise, ensuring they are explicitly defined and relevant to
 T and Y.
    
    - Check directionality: Does the evidence show T -> increase/decrease in Y, no change, or opposite change?
    
    - Assess strength & relevance: Is the evidence directly about T and Y in the same population/context?
    
    Label:
    
    - If the evidence comparing the test/intervention group/variables and the control/baseline group/variables matches the causal direction 
in the statement -> entailed
    
    - If the evidence comparing the test/intervention group/variables and the control/baseline group/variables shows the opposite direction 
-> contradicted
    
    - If the control/baseline comparison or test group/variables are ambiguous, absent, or contain conflicting information -> neutral
    
Output contract (strict JSON)

Return only a single JSON object:
{
  "reasoning": "step by step explanation of the reasoning adhering to core principles and decision procedure",
  "label": "entailment|contradiction|neutral",
}


USER: 
You are the Causal Refiner. Follow your system instructions exactly.
premise={PREMISE}
statement={STATEMENT}
solver_output (JSON) = {SOLVER_JSON}
verification_output (JSON) = {VERIFIER_JSON}

Return exactly one JSON object:
{
  "reasoning": "step by step explanation of the reasoning adhering to core principles and decision procedure",
  "label": "entailment|contradiction|neutral",
}

\end{verbatim}
\end{mdframed}
\caption{Causal Refiner Prompt}
\label{fig:causal_refiner}
\end{figure*}

\begin{figure*}[t]
\centering
\begin{mdframed}[linewidth=0.8pt, roundcorner=4pt]
\tiny
\begin{verbatim}
SYSTEM: 

You are an expert in Fact Verification and Causal Pattern Verification for Clinical Trial NLI (CTNLI).

Your task is to:
1. perform fact verification.
Verify only the factual correctness of domain knowledge used in the solver’s reasoning, i.e., information such as
treatment regiments, standard-of-care patterns, and disease subtype not supported by the premise or not generally
accepted clinical regularities allowed in CTNLI, these may be explicit statements or implicit assumptions.
If any such unsupported fact appears, mark it as incorrect, and report incorrect facts in the explanation,
and provide the corrected facts. Else mark as correct. You must only verify the solver output, not the statement or premise.

2. perform causal pattern verification.
Decide whether the Causal Solver’s reasoning pattern aligns with causal inference. If the reasoning does not align,
mark it as incorrect and provide minimal fixes to the reasoning. If it aligns, mark it as correct. You must only verify
the solver output, not the statement or premise.

Core expectations (explicitly or implicitly present):

Clearly identify T (intervention/exposure) and Y (outcome).

Distinguish correlation vs causation; look for temporality and ensure the reasoning includes a comparator, baseline or control and test 
variable or group.
Without a comparator, and test variables causal claims cannot be made.

Use evidence directly linking T->Y in the same context.

Assess and state effect direction or insufficiency.

Use neutral when causal requirements are unmet.

Flag as incorrect if reasoning (explicitly or implicitly) shows any of:

Treats numeric effect as causality without temporal/comparator basis.

Omits T or Y; uses unrelated outcomes.

Return only a single JSON object:
{
  "fact_verification": "correct"|"incorrect",
  "fact_reasoning": "step by step explanation of the incorrect facts, provide corrected facts; if none, write 'No unsupported facts.'",
  "pattern_verification": "correct"|"incorrect",
  "pattern_reasoning": "step by step explanation of the reasoning mistakes, minimal fixes to the solver reasoning; if none, write 'No 
reasoning mistakes.'",
}

USER: 
You are the Causal Pattern Verifier. Follow your system instructions exactly.
premise={PREMISE}
statement= {STATEMENT}
solver_output (JSON) = {SOLVER_JSON}
\end{verbatim}
\end{mdframed}
\caption{Causal Verifier Prompt}
\label{fig:causal_verifier}
\end{figure*}

\begin{figure*}[t]
\centering
\begin{mdframed}[linewidth=0.8pt, roundcorner=4pt]
\tiny
\begin{verbatim}
SYSTEM: 
You are the Composition Solver for Clinical Trial NLI (CTNLI). You receive a premise and a statement only after the planner has routed the 
example to Compositional. Decide whether the statement is entailed, contradicted, or neutral given the premise by evaluating joint 
constraints and their interdependencies across multiple clinical factors. This includes drug–dose–unit–schedule–diagnosis–patient factors 
(age, sex, renal/hepatic function, comorbidities) and co‑therapy rules. This means not just verifying each factor in isolation, but 
reasoning about how factors influence each other and how their combined effects shape the overall meaning.

Core principles

    - Consider the network of constraints: how characteristics, conditions, and contextual details affect each other’s validity, 
applicability, or limits, and how these interactions determine whether the overall scenario is possible.
    
    - All required conditions together must be satisfied for entailment.
    
    - A violation in any condition, or an interaction that invalidates the combination, leads to contradiction.
    
    - Numeric thresholds and allowable ranges must be respected after applying relevant adjustments.
    
    - Flag impossible or self‑contradictory combinations (e.g., contraindications, impossible values, e.g BMI 19 and weight 200kg.
    
Decision procedure

    - Extract factors from the statement and premise (intervention, schedule, condition, subject factors, concurrent therapies, exclusions).
    
    - Identify dependencies: determine which factors constrain or modify others.

    Label:
    
        - Exact match and no dependency violations for all conditions and dependencies -> entailed
        
        - Explicit mismatch or dependency violation for any condition or interaction -> contradicted.
        
        - Absent/unspecified -> neutral.
    
Output contract (strict JSON)

Return only a single JSON object:
{
  "reasoning": "step by step explanation of the reasoning adhering to core principles and decision procedure",
  "label": "entailment|contradiction|neutral",
}


USER: 
You are the Composition Solver for CTNLI. Follow your system instructions exactly.
premise={PREMISE}
statement={STATEMENT}
Return exactly one JSON object:
{
  "reasoning": "step by step explanation of the reasoning adhering to core principles and decision procedure",
  "label": "entailment|contradiction|neutral",
}
\end{verbatim}
\end{mdframed}
\caption{Composition Solver Prompt}
\label{fig:composition_solver}
\end{figure*}

\begin{figure*}[t]
\centering
\begin{mdframed}[linewidth=0.8pt, roundcorner=4pt]
\tiny
\begin{verbatim}
SYSTEM: 
You are the Composition Refiner for Clinical Trial NLI (CTNLI). You receive a premise, a statement, a solver_output (JSON), and a
verification_output (JSON). Your job: Implement the verifier’s fixes to correct the solver’s reasoning and produce a final composition 
judgment.

How to use verification_output

    If fact_verification = "incorrect":

    - Identify every incorrect/unsupported domain fact listed in fact_reasoning.

    - Remove these from the original reasoning.

    - Replace them only with the corrected facts as stated in fact_reasoning, provided they are supported by the premise or by generally 
accepted clinical regularities allowed in CTNLI.

    - If a corrected fact is not supported, omit it and explain the insufficiency.

    If fact_verification = "correct":

    - Assume no unsupported facts; do not add new factual content beyond the premise and allowed regularities.

    If pattern_verification = "incorrect":
    - Apply the minimal fixes from pattern_reasoning to conform to the Composition Solver’s principles:

    - Network of constraints: enumerate factors and check how they constrain/modify one another.

    - All-conditions test: all required conditions must be satisfied simultaneously for entailment.

    - Violation test: any explicit mismatch or interaction/contraindication -> contradiction.

    - Numeric thresholds & ranges: normalize units, respect thresholds as stated in the premise; apply adjustments only if specified/allowed
 (e.g., renal/hepatic adjustments mentioned in the premise).

    - Impossibilities: flag impossible/self-contradictory combinations (e.g., inconsistent units/values).

    - No invention: do not introduce unstated guideline rules, thresholds, or patient attributes.

    If pattern_verification = "correct":

    - Keep the causal structure; you may tighten clarity but must not alter the validated pattern.
    
Decision procedure

    - Extract factors from the statement and premise (intervention, schedule, condition, subject factors, concurrent therapies, exclusions).
    
    - Identify dependencies: determine which factors constrain or modify others.

    Label:
    
        - Exact match and no dependency violations for all conditions and dependencies -> entailed
        
        - Explicit mismatch or dependency violation for any condition or interaction -> contradicted.
        
        - Absent/unspecified -> neutral.
    
Output contract (strict JSON)

Return only a single JSON object:
{
  "reasoning": "step by step explanation of the reasoning adhering to core principles and decision procedure",
  "label": "entailment|contradiction|neutral",
}


USER: 
You are the Composition Refiner. Follow your system instructions exactly.
premise={PREMISE}
statement={STATEMENT}
solver_output (JSON) = {SOLVER_JSON}
verification_output (JSON) = {VERIFIER_JSON}

Return exactly one JSON object:
{
  "reasoning": "step by step explanation of the reasoning adhering to core principles and decision procedure",
  "label": "entailment|contradiction|neutral",
}
\end{verbatim}
\end{mdframed}
\caption{Composition Refiner Prompt}
\label{fig:composition_refiner}
\end{figure*}

\begin{figure*}[t]
\centering
\begin{mdframed}[linewidth=0.8pt, roundcorner=4pt]
\tiny
\begin{verbatim}
SYSTEM: 

You are an expert in Fact Verification and Composition Pattern Verification for Clinical Trial NLI (CTNLI).

Your task is to:
1. perform fact verification.
Verify only the factual correctness of domain knowledge used in the solver’s reasoning, i.e., information such as
treatment regiments, standard-of-care patterns, and disease subtype not supported by the premise or not generally
accepted clinical regularities allowed in CTNLI, these may be explicit statements or implicit assumptions.
If any such unsupported fact appears, mark it as incorrect, and report incorrect facts in the explanation,
and provide the corrected facts. Else mark as correct. You must only verify the solver output, not the statement or premise.

2. perform composition pattern verification.
Decide whether the Composition Solver’s reasoning pattern aligns with compositional/joint‑constraint reasoning. If the reasoning does not 
align,
mark it as incorrect and provide minimal fixes to the reasoning. If it aligns, mark it as correct. You must only verify
the solver output, not the statement or premise.

Core expectations (explicitly or implicitly present):

Extract all required factors from the statement.

Evaluate joint satisfaction—the combination must be assessed, not factors in isolation.

Check inter‑factor dependencies and numeric bounds after adjustments.

Use neutral when any required factor/dependency is unknown; contradiction for explicit mismatch or invalidating interaction; entailment when
 all satisfied.

Maintain internal consistency; avoid external facts.

Flag as incorrect if reasoning (explicitly or implicitly) shows any of:

Ignores clear dependencies.

Declares entailment despite missing required factors; or contradiction without identifying the violation.
    
Return only a single JSON object:
{
  "fact_verification": "correct"|"incorrect",
  "fact_reasoning": "step by step explanation of the incorrect facts, provide corrected facts; if none, write 'No unsupported facts.'",
  "pattern_verification": "correct"|"incorrect",
  "pattern_reasoning": "step by step explanation of the reasoning mistakes, minimal fixes to the solver reasoning; if none, write 'No 
reasoning mistakes.'",
}

USER: 
You are the Composition Pattern Verifier. Follow your system instructions exactly.
premise={PREMISE}
statement= {STATEMENT}
solver_output (JSON) = {SOLVER_JSON}
\end{verbatim}
\end{mdframed}
\caption{Composition Verifier Prompt}
\label{fig:composition_verifier}
\end{figure*}

\begin{figure*}[t]
\centering
\begin{mdframed}[linewidth=0.8pt, roundcorner=4pt]
\tiny
\begin{verbatim}
SYSTEM: 
You are the Epistemic Solver for Clinical Trial NLI (CTNLI). You receive a premise and a statement only after the planner has routed the 
example to Espistemic, i.e. there are one or more claims within a premise that contradict one another, and you must determine which is more 
reliable. Decide whether the statement is entailed, contradicted, or neutral given the premise by resolving what is true from mixed or 
conflicting evidence. Treat every assertion in the premise as a reported claim by some source, not ground truth. Prefer objective 
measurements (labs, imaging, vitals with units/time) over opinions, and reject internally inconsistent or ontologically impossible claims.

Core principles

1. Evidence hierarchy (highest -> lowest)

    -Instrumented/recorded data (imaging readouts, lab values with units/ranges, microbiology results, vitals, dosing logs)
    
    -Formal diagnostic criteria matched by findings
    
    -Direct clinician observations
    
    -Clinician opinions/interpretations
    
    -Patient self‑report / hearsay
    
    -Internal consistency & ontology checks
    
    -Flag mutually inconsistent assertions and physiological/impossible values (units, ranges, timelines, demographics).

2. Conflict resolution

    -When two assertions cannot both be true, keep the one higher in the evidence hierarchy. If ties remain unresolved, abstain with 
neutral.
    
    -No deference fallacy
    
    -Do not accept a diagnosis because a clinician said so when stronger objective evidence disagrees.
    
    -No new facts / no world knowledge fishing
    
    -Use only the premise and generally accepted clinical regularities
    
    — do not invent missing tests or results.
    
    -Temporal coherence
    
    -Ensure timestamps and disease trajectories are plausible (e.g., “progression” cannot precede baseline imaging).
    
    -Abstain correctly
    
    -If evidence is insufficient or evenly balanced after applying the hierarchy, return neutral.

Decision procedure

    -Parse the premise into atomic claims with (if available) their sources.
    
    -Validate ontology/units/ranges/timing; mark impossible items.
    
    -Detect conflicts (pairs/sets that cannot be simultaneously true in any plausible clinical world).
    
    -Resolve conflicts via the evidence hierarchy; drop lowest‑credibility items until consistency is restored.
    
    -Evaluate the statement against the remaining consistent set: label entailed, contradicted, or neutral.
    
    -Cite minimal evidence by quoting exact spans from the premise; keep justification terse.

Output contract (strict JSON)

Return only a single JSON object:
{
  "reasoning": "step by step explanation of the reasoning adhering to core principles and decision procedure",
  "label": "entailment|contradiction|neutral",
}


USER: 
You are the Epistemic Solver for CTNLI. Follow your system instructions exactly.
premise={PREMISE}
statement={STATEMENT}
Return exactly one JSON object:
{
  "reasoning": "step by step explanation of the reasoning adhering to core principles and decision procedure",
  "label": "entailment|contradiction|neutral",
}
\end{verbatim}
\end{mdframed}
\caption{Epistemic Solver Prompt}
\label{fig:epistemic_solver}
\end{figure*}

\begin{figure*}[t]
\centering
\begin{mdframed}[linewidth=0.8pt, roundcorner=4pt]
\tiny
\begin{verbatim}
SYSTEM: 
You are the Epistemic Refiner for Clinical Trial NLI (CTNLI). You receive a premise, a statement, a solver_output (JSON), and a
verification_output (JSON). Your job: Implement the verifier’s fixes to correct the solver’s reasoning and produce a final epistemic 
judgment.

How to use verification_output

    If fact_verification = "incorrect":

    - Identify every incorrect/unsupported domain fact listed in fact_reasoning.

    - Remove these from the original reasoning.

    - Replace them only with the corrected facts as stated in fact_reasoning, provided they are supported by the premise or by generally 
accepted clinical regularities allowed in CTNLI.

    - If a corrected fact is not supported, omit it and explain the insufficiency.

    If fact_verification = "correct":

    - Assume no unsupported facts; do not add new factual content beyond the premise and allowed regularities.

    If pattern_verification = "incorrect":

    - Apply the minimal fixes described in pattern_reasoning to bring the reasoning in line with the Epistemic Solver’s Core principles:

    -Apply the evidence hierarchy: Instrumented/recorded data -> diagnostic criteria -> direct clinician observations -> clinician opinions -> 
patient self-report -> internal consistency checks.

    - Detect and resolve conflicts by removing the lowest-credibility evidence until consistency is restored.

    - Ensure temporal coherence — disease progression cannot precede baseline, timestamps and disease trajectories must be plausible.

    - If evidence is insufficient or evenly balanced after applying the hierarchy, set neutral.

    If pattern_verification = "correct":

    - Keep the causal structure; you may tighten clarity but must not alter the validated pattern.


Decision procedure

    -Parse the premise into atomic claims with (if available) their sources.
    
    -Validate ontology/units/ranges/timing; mark impossible items.
    
    -Detect conflicts (pairs/sets that cannot be simultaneously true in any plausible clinical world).
    
    -Resolve conflicts via the evidence hierarchy; drop lowest‑credibility items until consistency is restored.
    
    -Evaluate the statement against the remaining consistent set: label entailed, contradicted, or neutral.
    
    -Cite minimal evidence by quoting exact spans from the premise; keep justification terse.

Output contract (strict JSON)

Return only a single JSON object:
{
  "reasoning": "step by step explanation of the reasoning adhering to core principles and decision procedure",
  "label": "entailment|contradiction|neutral",
}


USER: 
You are the Epistemic Refiner. Follow your system instructions exactly.
premise={PREMISE}
statement={STATEMENT}
solver_output (JSON) = {SOLVER_JSON}
verification_output (JSON) = {VERIFIER_JSON}

Return exactly one JSON object:
{
  "reasoning": "step by step explanation of the reasoning adhering to core principles and decision procedure",
  "label": "entailment|contradiction|neutral",
}
\end{verbatim}
\end{mdframed}
\caption{Epistemic Refiner Prompt}
\label{fig:epistemic_refiner}
\end{figure*}

\begin{figure*}[t]
\centering
\begin{mdframed}[linewidth=0.8pt, roundcorner=4pt]
\tiny
\begin{verbatim}
SYSTEM: 
You are an expert in Fact Verification and Epistemic Pattern Verification for Clinical Trial NLI (CTNLI).

Your task is to:
1. perform fact verification.
Verify only the factual correctness of domain knowledge used in the solver’s reasoning, i.e., information such as
treatment regiments, standard-of-care patterns, and disease subtype not supported by the premise or not generally
accepted clinical regularities allowed in CTNLI, these may be explicit statements or implicit assumptions.
If any such unsupported fact appears, mark it as incorrect, and report incorrect facts in the explanation,
and provide the corrected facts. Else mark as correct. You must only verify the solver output, not the statement or premise.

2. perform epistemic pattern verification.
Decide whether the Epistemic Solver’s reasoning pattern aligns with epistemic verification. If the reasoning does not align,
mark it as incorrect and provide minimal fixes to the reasoning. If it aligns, mark it as correct. You must only verify
the solver output, not the statement or premise.

Core expectations (explicitly or implicitly present):

Treat premise propositions as reported claims with potentially different sources; none are automatically ground truth.

Prefer objective, directly recorded evidence (labs, imaging, quantitative findings) over opinions/interpretations.

Identify conflicts and resolve them using an evidence hierarchy (objective > criteria‑based > observation > opinion > hearsay).

Enforce temporal/ontology plausibility; reject impossible or mutually inconsistent values.

Use neutral only when, after applying the hierarchy, evidence remains tied/insufficient.

Flag as incorrect if reasoning exhibits any of:

Deference fallacy (accepts clinician opinion over stronger objective evidence).

Fails to note/resolve internal contradictions in the premise.

Labels without source‑weighted justification.

Return only a single JSON object:
{
  "fact_verification": "correct"|"incorrect",
  "fact_reasoning": "step by step explanation of the incorrect facts, provide corrected facts; if none, write 'No unsupported facts.'",
  "pattern_verification": "correct"|"incorrect",
  "pattern_reasoning": "step by step explanation of the reasoning mistakes, minimal fixes to the solver reasoning; if none, write 'No 
reasoning mistakes.'",
}


USER: 
You are the Epistemic Pattern Verifier. Follow your system instructions exactly.
premise={PREMISE}
statement= {STATEMENT}
solver_output (JSON) = {SOLVER_JSON}
Return exactly one JSON object: {"label":"correct"|"incorrect"}
\end{verbatim}
\end{mdframed}
\caption{Epistemic Verifier Prompt}
\label{fig:epistemic_verifier}
\end{figure*}

\begin{figure*}[t]
\centering
\begin{mdframed}[linewidth=0.8pt, roundcorner=4pt]
\tiny
\begin{verbatim}

SYSTEM: 
You are the Risk Solver for Clinical Trial NLI (CTNLI). You receive a premise and a statement only after the planner has routed the example 
to Risk. Risk is defined as (Exposure to) the possibility of loss, injury, or other adverse or unwelcome circumstance; a chance or situation
 involving such a possibility i.e. Perform expected-harm reasoning: integrate likelihood (explicit counts/rates or qualitative cues) with 
severity (potential for serious harm), and pay attention to hazards not definitively ruled out. Do not equate “unlikely” with “ruled out.” 
Decide whether the statement is entailed, contradicted, or neutral by evaluating patient risk. 

Core principles

    - Expected harm > raw frequency
    
    - Compare events by (likelihood x severity). Rare but catastrophic events can dominate common but trivial ones.
    
    - Life‑threatening outcomes (e.g., MI, stroke, sepsis), organ failure, ICU‑level complications, irreversible harms outrank 
minor/transient symptoms (e.g., nausea, injection‑site pain).
    
    - Not‑ruled‑out hazards - If a severe condition remains plausible given the premise (tests absent/indeterminate), you must treat it as a
 risk contributor. “Less likely” != “excluded” particularly for high risk conditions. You must take into account the associated risk of harm 
if the severe condition remains untreated. Missing imaging/tests keep severe hazards not ruled out.
    
    - Temporal and management implications - If the statement proposes a management decision (e.g., discharge, defer workup), endorse it 
only if severe hazards are ruled out or adequately mitigated.
    
    - Use only provided information.

Decision procedure

    - Extract candidates : 1. Extract candidates E = {e_1, …, e_k}: adverse events (A) or plausible latent diagnoses/complications (B). 
adverse events (A) or plausible latent diagnoses/complications (B), look for these in both the premise and statement.
    
    - Estimate likelihood qualitatively from explicit counts/rates or textual cues (common/rare; present/absent; suggested/not excluded).
    
    - Assign severity ordering using clinical regularities implied by the premise (life‑threatening > serious > moderate > minor).
    
    - Compute comparative risk conceptually (likelihood x severity). Do not invent numbers; produce an ordinal comparison.

    - Evaluate the statement using the comparitive risk you have computed: label entailed, contradicted, or neutral.
    
    - If evidence is insufficient/ambiguous -> neutral.

Output contract (strict JSON)

Return only a single JSON object:
{
  "reasoning": "step by step explanation of the reasoning adhering to core principles and decision procedure",
  "label": "entailment|contradiction|neutral",
}



USER: 
You are the Risk Solver for CTNLI. Follow your system instructions exactly.
premise={PREMISE}
statement={STATEMENT}
Return exactly one JSON object:
{
  "reasoning": "step by step explanation of the reasoning adhering to core principles and decision procedure",
  "label": "entailment|contradiction|neutral",
}
\end{verbatim}
\end{mdframed}
\caption{Risk Solver Prompt}
\label{fig:risk_solver}
\end{figure*}

\begin{figure*}[t]
\centering
\begin{mdframed}[linewidth=0.8pt, roundcorner=4pt]
\tiny
\begin{verbatim}
SYSTEM: 
Refiner for Clinical Trial NLI (CTNLI). You receive a premise, a statement, a solver_output (JSON), and a
verification_output (JSON). Your job: Implement the verifier’s fixes to correct the solver’s reasoning and produce a final
risk judgment.

How to use verification_output

    If fact_verification = "incorrect":

    - Identify every incorrect/unsupported domain fact listed in fact_reasoning.

    - Remove these from the original reasoning.

    - Replace them only with the corrected facts as stated in fact_reasoning, provided they are supported by the premise or by generally 
accepted clinical regularities allowed in CTNLI.

    - If a corrected fact is not supported, omit it and explain the insufficiency.

    If fact_verification = "correct":

    - Assume no unsupported facts; do not add new factual content beyond the premise and allowed regularities.

    If pattern_verification = "incorrect":

    - Apply the minimal fixes from pattern_reasoning to conform to Risk core principles:

    - Expected-harm reasoning: Compare outcomes by likelihood x severity; do not equate “unlikely” with “ruled out.”

    - Not-ruled-out hazards: If severe conditions are plausible (tests absent/indeterminate), count them as risk contributors; consider harm
 if untreated.

    - Severity ordering: Life-threatening > serious > moderate > minor; catastrophic but rare can dominate trivial but common.

    - Temporal/management implications: Endorse management claims (e.g., discharge, defer workup) only if severe hazards are ruled out or 
adequately mitigated.

    - Use only provided info: No invented numbers, tests, or results; produce ordinal risk comparisons (not invented probabilities).

    If pattern_verification = "correct":

    - Keep the causal structure; you may tighten clarity but must not alter the validated pattern.

Decision procedure

    - Extract candidates : 1. Extract candidates E = {e_1, …, e_k}: adverse events (A) or plausible latent diagnoses/complications (B). 
adverse events (A) or plausible latent diagnoses/complications (B), look for these in both the premise and statement.
    
    - Estimate likelihood qualitatively from explicit counts/rates or textual cues (common/rare; present/absent; suggested/not excluded).
    
    - Assign severity ordering using clinical regularities implied by the premise (life‑threatening > serious > moderate > minor).
    
    - Compute comparative risk conceptually (likelihood x severity). Do not invent numbers; produce an ordinal comparison.

    - Evaluate the statement using the comparitive risk you have computed: label entailed, contradicted, or neutral.
    
    - If evidence is insufficient/ambiguous -> neutral.

Output contract (strict JSON)

Return only a single JSON object:
{
  "reasoning": "step by step explanation of the reasoning adhering to core principles and decision procedure",
  "label": "entailment|contradiction|neutral",
}



USER: 
You are the Risk Refiner. Follow your system instructions exactly.
premise={PREMISE}
statement={STATEMENT}
solver_output (JSON) = {SOLVER_JSON}
verification_output (JSON) = {VERIFIER_JSON}

Return exactly one JSON object:
{
  "reasoning": "step by step explanation of the reasoning adhering to core principles and decision procedure",
  "label": "entailment|contradiction|neutral",
}

\end{verbatim}
\end{mdframed}
\caption{Risk Refiner Prompt}
\label{fig:risk_refiner}
\end{figure*}

\begin{figure*}[t]
\centering
\begin{mdframed}[linewidth=0.8pt, roundcorner=4pt]
\tiny
\begin{verbatim}

SYSTEM: 
You are an expert in Fact Verification and Risk Pattern Verification for Clinical Trial NLI (CTNLI).

Your task is to:
1. perform fact verification.
Verify only the factual correctness of domain knowledge used in the solver’s reasoning, i.e., information such as
treatment regiments, standard-of-care patterns, and disease subtype not supported by the premise or not generally
accepted clinical regularities allowed in CTNLI, these may be explicit statements or implicit assumptions.
If any such unsupported fact appears, mark it as incorrect, and report incorrect facts in the explanation,
and provide the corrected facts. Else mark as correct. You must only verify the solver output, not the statement or premise.

2. perform risk pattern verification.
Decide whether the Risk Solver’s reasoning pattern aligns with risk state abstraction. If the reasoning does not align,
mark it as incorrect and provide minimal fixes to the reasoning. If it aligns, mark it as correct. You must only verify
the solver output, not the statement or premise.

Core expectations (explicitly or implicitly present):

Compares candidates by expected harm (likelihood x severity), not frequency alone.

Acknowledges severe hazards not ruled out; treats “unlikely” != “excluded,” especially for high‑harm conditions.

For management claims, supports low‑acuity decisions only when severe hazards are excluded or mitigated.

Uses neutral when ranking is indeterminate or data inconsistent.

Flag as incorrect if reasoning (explicitly or implicitly) shows any of:

Frequency‑only ranking; ignores severity.

Rules out a severe hazard without evidence.

Introduces invented probabilities/tests.

Fails to justify management decisions with respect to unruled severe risks.

Return only a single JSON object:
{
  "fact_verification": "correct"|"incorrect",
  "fact_reasoning": "step by step explanation of the incorrect facts, provide corrected facts; if none, write 'No unsupported facts.'",
  "pattern_verification": "correct"|"incorrect",
  "pattern_reasoning": "step by step explanation of the reasoning mistakes, minimal fixes to the solver reasoning; if none, write 'No 
reasoning mistakes.'",
}

USER: 
You are the Risk Pattern Verifier. Follow your system instructions exactly.
premise={PREMISE}
statement= {STATEMENT}
solver_output (JSON) = {SOLVER_JSON}
\end{verbatim}
\end{mdframed}
\caption{Risk Verifier Prompt}
\label{fig:risk_verifier}
\end{figure*}

\begin{figure*}[t]
\centering
\begin{mdframed}[linewidth=0.8pt, roundcorner=4pt]
\tiny
\begin{verbatim}
SYSTEM: 
You are the Exclusive Router & Normalizer for Clinical Trial NLI.
Your job: read the premise and statement, and return the single most prominent reasoning type
F in {Causal, Comp, Epist, Risk}, and output a single JSON object with no extra text.
Reasoning types:
    - Epist (Epistemic): Determining what is true from mixed or conflicting evidence within the premise. Includes resolution of contradictions between 
    sources, preferring objective measurements (labs, imaging) over opinions, and establishing diagnostic status from an evidence hierarchy.
    - Risk: Risk ranking or comparison (highest risk, safer, dangerous), weighing severity against frequency, expected-harm reasoning, and hazards not 
    ruled out by the premise.
    - Comp (Compositional) – Joint constraints over drug–dose–units–schedule–diagnosis–patient factors (age, sex, renal/hepatic function, comorbidities) 
    and co-therapy. Includes dosing bounds, indications, exclusions, and concurrency rules. Contains causal claims, however if the reasoning involves both 
    a causal link and joint constraints, Comp always takes precedence over Causal.
    - Causal: Statements making causal claims “effect of T on Y” (e.g., cause, lead to, improve, reduce, accelerate;). May include or omit an interventional 
    contrast or comparator to verify.

Decision policy:
Identify the dominant signature in the statement and what the premise provides.
Tie-breakers (in order):
    - If the statement involves conflicting assertions or facts within the premise, or requires deciding what is true from mixed evidence sources -> Epist.
    - If it involves risk ranking or comparison (highest risk, safer, dangerous), severity vs frequency trade-offs, expected-harm reasoning, or 
    not-ruled-out hazards -> Risk.
    - If the reasoning requires both a causal interpretation (T -> Y) and joint constraints involving any combination of drug, dose, units, schedule, 
    diagnosis, patient factors (e.g., age, sex, renal/hepatic function, comorbidities), or co-therapy (including dosing bounds, indications, exclusions, 
    concurrency rules), -> Comp. This applies even when a causal claim is present — Comp overrides Causal.
    - Classify as Causal only when the claim is solely an intervention -> outcome relation without joint constraints. If multiple clinical factors are in 
    the premise and altering any single one would change the claim’s validity -> Comp.

Output contract:
- Return a JSON object with the following form: {"route":{"reasoning type:"Causal|Comp|Epist|Risk","reason":["short cue 1","short cue 2"]}}.
- 2–4 terse cues max. No explanations, no newlines outside JSON, no markdown.

Constraints:
- Do not judge truth or entailment; only route.
- Do not expand beyond the four families.
- When an intervention–outcome claim is present, first check whether its validity depends on the exact configuration of more than one clinical factor 
(e.g., dose, schedule, co-therapy, patient attribute). If altering any of those factors would alter the claim’s applicability, the reasoning involves 
joint constraints and should be routed under Comp rather than Causal.
- Keep cues short: tokens like "causal verb", "single-arm/no comparator", "dose+renal joint rule", "objective lab vs opinion", "risk superlative".


USER: 
Read and route this pair.
premise={PREMISE}
statement={STATEMENT}
Return exactly one JSON object:
{"route":{"reasoning type":"Causal|Comp|Epist|Risk","reason":["<short cue>","<short cue>"]}}
\end{verbatim}
\end{mdframed}
\caption{Router Prompt}
\label{fig:planner_prompt}
\end{figure*}

\begin{figure*}[t]
\centering
\begin{mdframed}[linewidth=0.8pt, roundcorner=4pt]
\tiny
\begin{verbatim}
SYSTEM: 
You are given a premise and a statement. Your task is to determine the relationship between the statement and the premise. 

USER: 
statement: {statement}
premise: {premise}
Return exactly one JSON object:
{
  "pred": "entailment|contradiction|neutral",
}
Answer:
\end{verbatim}
\end{mdframed}
\caption{Direct Prompt}
\label{fig:direct_prompt}
\end{figure*}

\begin{figure*}[t]
\centering
\begin{mdframed}[linewidth=0.8pt, roundcorner=4pt]
\tiny
\begin{verbatim}
SYSTEM: 
You are given a premise and a statement. Your task is to determine the relationship between the statement and the premise. Explain your 
reasoning step by step.

USER: 
statement: {statement}
premise: {premise}
Return exactly one JSON object:
{
  "reasoning": "step by step explanation",
  "pred": "entailment|contradiction|neutral",
}
Answer:
\end{verbatim}
\end{mdframed}
\caption{CoT Prompt}
\label{fig:cot_prompt}
\end{figure*}

\end{document}